\documentclass[11pt,a4paper]{article}
\usepackage{times,latexsym}
\usepackage{url}
\usepackage[T1]{fontenc}

\usepackage[acceptedWithA]{tacl2021v1}
\usepackage{xspace,mfirstuc,tabulary}

\usepackage{amsmath}
\DeclareMathOperator*{\argmax}{arg\,max}
\usepackage{multirow}
\usepackage{graphicx}
\usepackage{times}
\usepackage{latexsym}
\usepackage{xcolor}
\usepackage{graphicx}
\usepackage{amsmath, bm}
\usepackage{amsfonts}
\usepackage{multirow}
\usepackage{cleveref}
\usepackage{booktabs}
\usepackage{xcolor}
\usepackage{enumitem}
\usepackage{amssymb}
\usepackage{pifont}
\usepackage{arydshln}
\usepackage{svg}
\usepackage{tabularx}
\usepackage{CJK}
\usepackage{array}
\usepackage{makecell}

\usepackage{pifont}       % \ding{xx}
\usepackage{bbding}       % \Checkmark,\XSolid,... (需要和pifont宏包共同使用)
\usepackage{fontawesome}
\usepackage{subfigure}
\usepackage[normalem]{ulem}
\usepackage[hang,flushmargin]{footmisc}
\newcommand\cmark {\textcolor{green}{\ding{52}}}
\newcommand\xmark {\textcolor{red}{\ding{55}}}

\newif\iftaclinstructions
\taclinstructionsfalse % AUTHORS: do NOT set this to true
\iftaclinstructions
\renewcommand{\confidential}{}
\renewcommand{\anonsubtext}{(No author info supplied here, for consistency with
TACL-submission anonymization requirements)}
\newcommand{\instr}
\fi

\iftaclpubformat % this "if" is set by the choice of options

\else

\fi

\newcommand{\improvement}{6.07}
\newcommand{\pb}{$P_b$}
\title{An Energy-based Model for Word-level AutoCompletion \\
in Computer-aided Translation}

\author{
    Cheng Yang$^{1}$\Thanks{\hspace{-1mm}Work done during internship at Tencent AI Lab.}
    \quad
    Guoping Huang$^{2}$
    \quad
    Mo Yu$^{3}$
    \quad
    Zhirui Zhang$^{2}$
    \quad
    Siheng Li$^{1}$
    \\
    \textbf{Mingming Yang}$^{2}$
    \quad
    \textbf{Shuming Shi}$^{2}$
    \quad
    \textbf{Yujiu Yang}$^{1}$\Thanks{\hspace{-1mm}Corresponding Authors.}
    \quad
    \textbf{Lemao Liu}$^{2}$\footnotemark[2]
  \\
  $^{1}$Tsinghua Shenzhen International Graduate School, Tsinghua University, China
  \\
  $^{2}$Tencent AI Lab \quad $^{3}$WeChat AI, Tencent
  \\
  \texttt{yangc21@mails.tsinghua.edu.cn},
  \texttt{yang.yujiu@sz.tsinghua.edu.cn}, 
  \\
  \texttt{\{donkeyhuang,moyumyu,jackzrzhang,shanemmyang,}\\
  \texttt{shumingshi,redmondliu\}@tencent.com}
}

\date{}

\begin{document}
\maketitle
\begin{abstract}
Word-level AutoCompletion~(WLAC) is a rewarding yet challenging task in Computer-aided Translation.
Existing work addresses this task through a classification model based on a neural network that maps the hidden vector of the input context into its corresponding label (i.e., the candidate target word is treated as a label).
Since the context hidden vector itself does not take the label into account and it is projected to the label through a linear classifier, the model can not sufficiently leverage valuable information from the source sentence as verified in our experiments, which eventually hinders its overall performance.
To alleviate this issue, this work proposes an energy-based model for WLAC, which enables the context hidden vector to capture crucial information from the source sentence.
Unfortunately, training and inference suffer from efficiency and effectiveness challenges,
thereby we employ three simple yet effective strategies to put our model into practice.
Experiments on four standard benchmarks demonstrate that our reranking-based approach achieves substantial improvements (about 6.07\%) over the previous state-of-the-art model.
Further analyses show that each strategy of our approach contributes to the final performance.\footnote{Our
codes are available at \url{https://github.com/yc1999/energy_wlac}}\looseness=-1
\end{abstract}

\section{Introduction}
\label{sec:1}

\begin{figure}[t]
    \centering
    \includegraphics[width=\columnwidth]{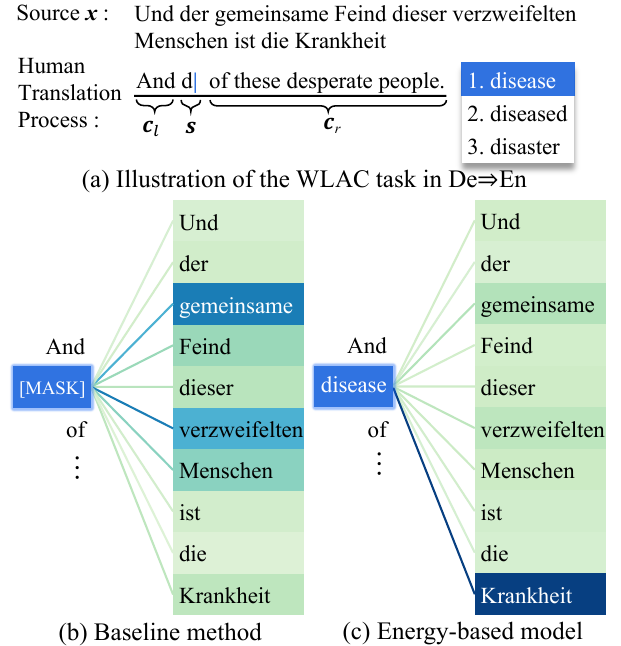}
    \caption{(a) Illustration of the WLAC task in De$\Rightarrow$En. Suppose that a user has input a source sentence $\boldsymbol{x}$, partial translations ($\boldsymbol{c}_l$, $\boldsymbol{c}_r$) and is now typing some characters ($\boldsymbol{s}$). A well-trained WLAC model is expected to suggest ``\emph{disease}'' to complete $\boldsymbol{s}$. The expected translation for $\boldsymbol{x}$ is ``And disease is the common enemy of these desperate people.'' (b) Attention weights from ``\texttt{[MASK]}'' to words in $\boldsymbol{x}$ of the baseline method. (c) Attention weights from ``\emph{disease}'' to words in $\boldsymbol{x}$ of our energy-based model. (Color intensity reflects the strength of attention weights.)
    }
    \label{fig:intro}
\end{figure}

Computer-aided Translation (CAT)
\citep{DBLP:journals/coling/BarrachinaBCCCKLNTVV09,DBLP:conf/emnlp/SantyDCB19,DBLP:journals/corr/abs-2105-13072},
which enables the leveraging of machine translation systems \citep{DBLP:journals/corr/BahdanauCB14,DBLP:conf/nips/VaswaniSPUJGKP17} to improve the efficiency of the human translation process,
has seen increasing interest in recent years.
In this work, we study a crucial yet challenging task in CAT: \textbf{W}ord-\textbf{L}evel \textbf{A}uto\textbf{C}ompletion (WLAC) \citep{DBLP:conf/acl/LiLHS20}, which aims at yielding word-level suggestions based on context pieces provided by human (Figure~\ref{fig:intro}(a)).\looseness=-1

Previous research includes statistical methods \citep{DBLP:conf/ijcai/HuangZZZ15} and neural methods \citep{DBLP:conf/emnlp/SantyDCB19,DBLP:conf/acl/LiLHS20}.
With the help of word alignment toolkits \citep{DBLP:journals/coling/OchN03,DBLP:conf/naacl/DyerCS13}, statistical approaches build a translation table and use it to predict the target word.
More recently, \citet{DBLP:conf/acl/LiLHS20} use a Transformer-based classification model,
which firstly encodes the input context to a hidden vector and then maps the hidden vector into the candidate target word through a linear classifier.
This strong baseline method achieves the state-of-the-art (SOTA) performance.\looseness=-1

In the aforementioned classification paradigm, the hidden vector of the input context inherently does not take the candidate target word into consideration.
As a result, it may not effectively leverage valuable information carried by the candidate target word when occurring in the input context, as shown in Figure~\ref{fig:intro}(b). Specifically,
given the input context and human typed characters ``\emph{d}'', the user may tend to type ``\emph{disease}'' (``\emph{Krankheit}'' in German). However, through visualizing attention weights, it shows that the baseline method captures more information from ``\emph{gemeinsame}'' and ``\emph{verzweifelten}'' than that from the most informative word ``\emph{Krankheit}'' in the source side, which may underestimate the model score of the ground-truth word ``\emph{disease}'' and thereby leads to incorrect prediction.\looseness=-1
 
To alleviate the above issue, we formalize the WLAC task with an \textit{energy-based model}~\cite{DBLP:conf/nips/RanzatoPCL06,lecun2006tutorial} based on Transformer, where the hidden vector is defined on top of both the candidate target word and the input context through a deep energy function.
Furthermore, with the help of deep neural networks, the energy-based function is expected to capture sufficient information for each candidate target word through the attention mechanism.
In this way, the energy function is able to capture informative context (i.e., ``\emph{Krankheit}'') to evaluate the target word (i.e., ``\emph{disease}''), and thereby the score from the energy-based model is more reliable, as shown in Figure~\ref{fig:intro}(c).\looseness=-1

Unfortunately, training and inference with the energy-based model suffer from efficiency and effectiveness challenges due to the normalization term in the model.
To alleviate the effect of these barriers, we systematically incorporate three simple yet effective strategies inspired by previous studies: (1)~a \textit{negative sampling} method for efficient training~\citep{DBLP:conf/emnlp/MaC18,DBLP:conf/emnlp/LiTWFZY19,DBLP:journals/corr/abs-2206-00212}, (2)~a \textit{reranking} paradigm as an approximate proxy for efficient inference~\citep{DBLP:conf/naacl/ShenSO04,DBLP:journals/corr/abs-1901-04085,DBLP:conf/acl/BhattacharyyaRN20} and (3) a \textit{pre-training} method for effective training~\cite{DBLP:conf/acl/0001AR20}.
Experiments on four standard benchmarks demonstrate that the energy-based model is indeed better at capturing informative signals for the prediction of a candidate target word and thereby yields substantial improvements over strong baselines. 

To sum up, our contribution is three-fold:
\begin{enumerate}[itemsep=0.2pt]
    \item We point out that the previous SOTA model for the WLAC task suffers from an issue, i.e., it can not sufficiently leverage the valuable information from the source sentence for word prediction. 
    \item We propose an energy-based model to alleviate this issue and we employ three simple yet effective strategies to put it into practice. 
    \item We comprehensively evaluate our approach on four benchmarks, and our approach achieves substantial improvements (about \improvement\%) over the previous SOTA model.
\end{enumerate}

\section{Preliminary}
In this section, we review the setting of the WLAC task and introduce the state-of-the-art baseline method, which will be reused in Section~\ref{sec:3}.\looseness=-1

\subsection{WLAC Task}
\label{sec:2.1}
\paragraph{Notations} Let $\boldsymbol{x} = (x_1, x_2, \dots , x_T )$ be a source sentence, $\boldsymbol{s} = ( s_1, s_2, \dots , s_k )$ be a sequence of human typed characters and $\boldsymbol{c} = ( \boldsymbol{c}_l, \boldsymbol{c}_r )$ be translation context where $ \boldsymbol{c}_l = ( c_{l,1}, c_{l,2}, \dots, c_{l,m} )$ and $ \boldsymbol{c}_r = ( c_{r,1}, c_{r,2}, \dots, c_{r,n} )$~.
$\boldsymbol{c}_{l}$ and $\boldsymbol{c}_{r}$ are on the left and right-hand side of $\boldsymbol{s}$, respectively. Figure~\ref{fig:intro}(a) illustrates the examples for $\boldsymbol{x}$, $\boldsymbol{c}_l$, $\boldsymbol{c}_r$, and $\boldsymbol{s}$. 

\paragraph{Task Definition}
Given the input tuple $(\boldsymbol{x}, \boldsymbol{c}, \boldsymbol{s})$, the \textbf{WLAC} task aims at predicting the target word $w$, which starts with $\boldsymbol{s}$ and is the most appropriate to be placed between $\boldsymbol{c}_l$ and $\boldsymbol{c}_r$~\cite{DBLP:conf/acl/LiLHS20}.\looseness=-1
In partial translation consisting of $\boldsymbol{c}_l$, $w$ and $\boldsymbol{c}_r$, $w$ is not necessary to be consecutive to $c_{l,m}$ and $c_{r,1}$. Figure~\ref{fig:intro}(a) gives an illustrative example.
To be more general in real-world scenarios, the WLAC task further assumes that $\boldsymbol{c}_l$ and $\boldsymbol{c}_r$ can be empty, which leads to following four translation context types:\looseness=-1
\begin{itemize}[wide=1.0\parindent,noitemsep, topsep=2pt]
    \item Zero-context: both $\boldsymbol{c}_l$ and $\boldsymbol{c}_r$ are empty;
    \item Prefix: $\boldsymbol{c}_r$ is empty;
    \item Suffix: $\boldsymbol{c}_l$ is empty;
    \item Bi-context: both $\boldsymbol{c}_l$ and $\boldsymbol{c}_r$ are not empty.
\end{itemize}

It is noteworthy that context types described above are general and encompass context of several conventional translation scenarios, such as prefix-decoding for left-to-right interactive machine translation~(IMT)~\citep{DBLP:conf/amta/KnowlesK16} and post-editing~\citep{DBLP:conf/acl/LeeAPJ21,DBLP:conf/emnlp/YangMZL022}.
To elaborate, in prefix-decoding, the context falls into the special case of prefix, where $\boldsymbol{c}_r$ is empty and $\boldsymbol{c}_l$ is consecutive to $w$.
In post-editing, the context corresponds to the special case of bi-context, where both $\boldsymbol{c}_l$ and $\boldsymbol{c}_r$ are consecutive to $w$.\looseness=-1

\subsection{Baseline Method}
\label{sec:2.2}
\citet{DBLP:conf/acl/LiLHS20} cast WLAC as a word prediction task.
Generally, they decompose the WLAC task into two steps: (1) Model the distribution of the target word $w$ using $\boldsymbol{x}$ and  $\boldsymbol{c}$ via a \textbf{Word Prediction Model} (WPM); (2) Predict the most appropriate word $\hat{w}$ which starts with $\boldsymbol{s}$ according to the conditional distribution.
Their method achieves state-of-the-art performance.

A baseline WPM is defined by Transformer architecture~\citep{DBLP:conf/nips/VaswaniSPUJGKP17} for NMT.
Specifically, it first uses a placeholder \texttt{[MASK]} to represent the position of the target word $w$ and put it between $\boldsymbol{c}_l$ and $\boldsymbol{c}_r$.
Ultimately, it uses the representation of \texttt{[MASK]} defined through Transformer to predict the target word. 
Figure~\ref{fig:framework}(a) shows the model architecture of the baseline WPM.
Formally, the conditional probability distribution of the target word $w$ is:\looseness-1
\begin{equation}
    P_b(w\mid\boldsymbol{x},\boldsymbol{c};\Theta) = \mathrm{softmax} (\mathbf{M} \mathbf{h}_{\mathrm{[MASK]}}^{\top})[w]
    \label{eq:disc}
\end{equation}
\noindent where $ \mathbf{h}_{ \mathrm{[MASK]} } $ is the dense representation of \texttt{[MASK]}, $\mathbf{M}$ represents the learnable embedding matrix, and $[w]$ denotes taking the component with respect to the index $w$. In the following sections, we use $P_{b}$ to denote the baseline WPM.

Then during the inference stage, \pb~tries to pick up the best $w$ according to the following equation:
\begin{align}
     & \argmax_{w\in \mathcal{V}(\boldsymbol{s})} P_b(w\mid\boldsymbol{x},\boldsymbol{c};\Theta)\notag  \\ 
     = & \argmax_{w\in \mathcal{V}(\boldsymbol{s})} \mathbf{M}[w] \mathbf{h}_{\mathrm{[MASK]}}^{\top} \label{eq:base-infer}
\end{align}
\noindent where $\mathcal{V}(\boldsymbol{s})$ denotes a set of candidate words that start with $\boldsymbol{s}$, and $\mathbf{M}[w]$ is the word embedding vector of $w$.   Note that $\mathbf{h}_{\mathrm{[MASK]}}$ is independent of $w$, and $\mathbf{M} \mathbf{h}_{\mathrm{[MASK]}}^{\top}$ can be efficiently computed with GPU in parallel. Therefore, $\argmax$ in Equation~\eqref{eq:base-infer} can be computed exactly. 
\section{Energy-based Model}
\label{sec:3}

\subsection{Motivation}
As shown in Equation~\eqref{eq:base-infer} in Section~\ref{sec:2.2},
the baseline WPM essentially maps the hidden vector of the input context~(i.e., $\mathbf{h}_{\mathrm{[MASK]}}$) into the candidate target word to predict the most appropriate target word for \texttt{[MASK]}.
Furthermore, according to the model architecture of the baseline WPM, the context hidden vector $\mathbf{h}_{\mathrm{[MASK]}}$ does not take the candidate target word into consideration~\cite{liu-etal-2016-neural,li-etal-2018-target}.
Therefore, it might be difficult for $\mathbf{h}_{\mathrm{[MASK]}}$ to make full use of sufficient information from the source side for accurately predicting the ground-truth target word. 
Intuitively, the above issue for the baseline WPM in Equation~\eqref{eq:disc} can be demonstrated from the example in Figure~\ref{fig:intro}(b), where we use attention weights to visualize source words which are mostly used in $\mathbf{h}_{\mathrm{[MASK]}}$
\footnote{In our preliminary experiments, we also employed other methods to attribute source words that are mostly used (e.g., the prediction difference method~\citep{DBLP:conf/acl/LiLLMS19}).
The conclusions drawn from these alternative methods align closely with those obtained using attention weights.
This suggests that, in the context of the WLAC task, the model's utilization of source-side information can be consistently reflected through various effective attribution methods.
In this paper, we opt to utilize attention weights for easier description.}. 
From this figure, it shows that $\mathbf{h}_{\mathrm{[MASK]}}$ uses more information from ``\emph{gemeinsame}'' and ``\emph{verzweifelten}'' than that from ``\emph{Krankheit}''. Therefore, such a model may underestimate the score for the ground-truth word ``\emph{disease}'', which aligns to ``\emph{Krankheit}'' on the source side. Consequently, the baseline WPM may not successfully predict the ground-truth word, leading to sub-optimal performance.\looseness=-1

In response to the above issue, this paper proposes an energy-based model which enables defining the hidden vector on top of both the candidate target word and the input context through an energy function.
Our intuition is that with the help of deep neural networks~(e.g., attention networks),
the energy function is expected to capture more valuable information from the source sentence, which makes the model score more reliable to evaluate contributions for $w$.

\begin{figure*}
    \centering
    \includegraphics[width=\linewidth]{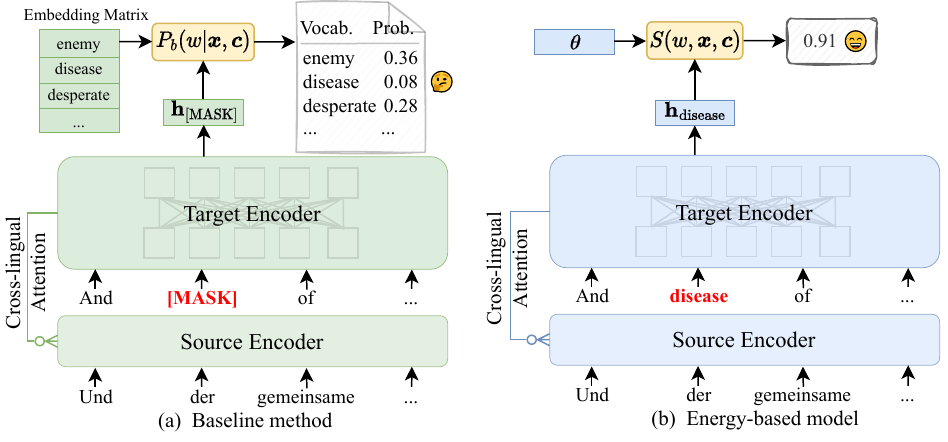}
    \caption{The comparison between the network architectures for the baseline method WPM (a) and the energy-based model (b). In the baseline model, $\mathbf{h}_\textrm{[MASK]}$ does not capture the information from ``\emph{disease}'' whereas $\mathbf{h}_\textrm{[disease]}$ does in the energy-based model. Note that ``Target Encoder'' is a variant of the Transformer decoder which can capture bidirectional information on the target side.}
    \label{fig:framework}
\end{figure*}

\subsection{Model Definition}
Formally, given $\boldsymbol{x}$ and $\boldsymbol{c}$, we employ an energy-based model to define the word prediction model as follows:
\begin{equation}
    P(w\mid \boldsymbol{x}, \boldsymbol{c}; \Theta) = \frac{\exp( S(w, \boldsymbol{x}, \boldsymbol{c}))}{Z(\boldsymbol{x}, \boldsymbol{c})}
    \label{eq:energy}
\end{equation}
with $$Z(\boldsymbol{x}, \boldsymbol{c})=\sum_w \exp( S(w, \boldsymbol{x}, \boldsymbol{c}))$$
\noindent where $S(w, \boldsymbol{x}, \boldsymbol{c})$ is an energy function taking a real value and $Z(\boldsymbol{x}, \boldsymbol{c})$ is the normalization term. 

The energy-based model in Equation~\eqref{eq:energy} is very general, because the energy function $S(w, \boldsymbol{x}, \boldsymbol{c})$ can be any function. For example, as a special case, if we set $S(w, \boldsymbol{x}, \boldsymbol{c})=P_b(w|\boldsymbol{x}, \boldsymbol{c})$, the energy-based model is then reduced to Equation~\eqref{eq:disc} because the normalization term is 1.
Since this paper aims to alleviate the insufficient usage of source sentence information for $P_b$,
it seeks another definition of the energy function to define the hidden vector on top of both the candidate target word~$w$ and the input context~($\boldsymbol{x}, \boldsymbol{c}$).

Theoretically, there are many ways to define the energy function $S(w,\boldsymbol{x}, \boldsymbol{c})$. 
In this paper, in practice, we adopt the way to define $S(w,\boldsymbol{x}, \boldsymbol{c})$ very similar to $P_b$ in model architecture with minimal modifications and almost the same number of parameters as $P_b$.
As a result, it could indicate that the potential improvement derived from the energy-based model is not significantly attributed to the complex model architecture of $S(w,\boldsymbol{x}, \boldsymbol{c})$, but rather to define the hidden vector on top of both the candidate target word~$w$ and the input context~($\boldsymbol{x}, \boldsymbol{c}$).

Specifically, the energy function $S$ adopts the similar Transformer architecture as $P_b$. $S$ differs from $P_b$ only in two aspects:
first, we replace the embedding matrix with a binary classifier. The binary classifier is defined by a parameterized weight vector and brings only a small number of parameters; second, in particular, the candidate target word $w$ is fed into the Transformer, then it is used as the query in the attention mechanism with $(\boldsymbol{x}, \boldsymbol{c})$.
With the help of deep neural networks, $S$ is expected to capture sufficient information for $w$ through the attention mechanism. 
Formally, the energy function is defined as follows: $$S(w, \boldsymbol{x}, \boldsymbol{c})=\mathrm{Sigmoid}\big(\theta\cdot \mathbf{h}(w, \boldsymbol{x}, \boldsymbol{c})^{\top}\big)$$
where $ \mathbf{h}$ is the dense representation vector of $w$ accompanied with $\boldsymbol{x}$ and $\boldsymbol{c}$, and $\theta$ is a learnable weight vector. The architecture of the energy function is illustrated in Figure~\ref{fig:framework}(b).

We believe that the energy function $S$ can adequately exploit contextual information from $(\boldsymbol{x},\boldsymbol{c})$.
This belief is exemplified in Figure~\ref{fig:intro}(c).\footnote{Note that this example is not cherry-picked and more quantitative analyses will be shown in the later experiments.}
In this figure, after visualizing attention weights to source words, the energy function $S$ is able to capture more information from ``\emph{Krankheit}'' to evaluate the target word ``\emph{disease}''.
Thereby $S(\mathrm{disease}, \boldsymbol{x},\boldsymbol{c})$ is more reliable than baseline score $P_b(\mathrm{disease}|\boldsymbol{x},\boldsymbol{c})$, which inadequately make use of the signal from ``\emph{Krankheit}'' as shown in Figure~\ref{fig:intro}(b).

\subsection{Challenges}
\label{sec:3.3}
However, it is far from trivial to make the energy-based model achieve the effect as shown in Figure~\ref{fig:intro}(c) and further deliver excellent performance on the WLAC task due to the following efficiency and effectiveness challenges.

\paragraph{Efficiency} The first challenge is the efficiency in both training and inference.
During training, maximizing the log-likelihood for Equation~\eqref{eq:energy} needs to calculate the value of the normalization term.
During inference, it needs to enumerate all candidate words from vocabulary $\mathcal{V}$.
Unfortunately, the energy function $S$ sacrifices the parallel computation for all $w\in \mathcal{V}$:
one has to feed all candidate target words to the network architecture independently for each $w$. 
However, since $\mathcal{V}$ is too large, such exhaustive computation is infeasible in practice. 
Consequently, this makes both training and inference challenging for the energy-based model.\looseness=-1

\paragraph{Effectiveness} Second, in our preliminary experiments, optimizing the energy-based model from scratch does not work well, and its final performance is significantly worse than the baseline $P_b$. One possible reason is that it is more difficult to train the energy-based model. Training the energy-based model involves an approximate method to shrink the subset for the normalization term, and this may induce a risk that the informative negative examples are excluded in the shrunk subset~\citep{DBLP:conf/emnlp/MaC18,DBLP:journals/corr/abs-2206-00212}. 
Therefore, it is easy to get trapped in local optimization when training the energy-based model from scratch.

\section{Training and Inference}
To relieve the aforementioned challenges, we systematically employ three simple yet effective methods inspired by previous studies.
First, we employ negative sampling to address the normalization computation during the training~\citep{DBLP:conf/emnlp/MaC18,DBLP:conf/emnlp/LiTWFZY19,DBLP:journals/corr/abs-2206-00212};   similarly, during the inference, we adopt a reranking paradigm, where the energy-based model is used as a reranker over a small subset of candidates~\citep{DBLP:conf/naacl/ShenSO04,DBLP:journals/corr/abs-1901-04085,DBLP:conf/acl/BhattacharyyaRN20}.
Moreover, we harness a conditional mask bilingual language modeling pre-training strategy for parameter initialization~\cite{DBLP:conf/acl/0001AR20}.

\subsection{Efficient Training and Inference}

\paragraph{Efficient Training via Negative Sampling}
As described in Section~\ref{sec:3.3}, it is infeasible to calculate the normalization term in an exact way.  To optimize the parameter $\Theta$ for the energy-based model in Equation~\eqref{eq:energy}, we instead use the negative sampling method to approximate the normalization term $Z(w, \boldsymbol{x}, \boldsymbol{c}; \Theta)$, and then we maximize the following objective function:
\begin{align}
   &  w_i \sim \hat{P} \textrm{ for } i\in [1, K] \label{eq:sample} \\
   &  S(w, \boldsymbol{x}, \boldsymbol{c}; \Theta) - 
     \log \big[ \sum_{i=1}^K \exp S(w_i, \boldsymbol{x}, \boldsymbol{c}; \Theta) \big] \notag
\end{align}
\noindent where $\hat{P}$ is a predefined and parameter-free distribution over the vocabulary $\mathcal{V}$ and $w_i\sim \hat{P}$ denotes sampling from the distribution $\hat{P}$. Note that if we consider all $w_i\in \mathcal{V}$, then the above objective function is equivalent to the likelihood function for the energy-based model in Equation~\eqref{eq:energy}.

In this paper, we try different settings for $\hat{P}$. 
As the first setting, $\hat{P}$ is defined by the uniform distribution over $\mathcal{V}$. Although sampling from this distribution is efficient and even does not introduce extra computation, it can not ensure the hard negatives are sampled with a high probability. Thus it is not promising to speed up the convergence in our experiments. Hence, as the second setting, $\hat{P}$ is instantiated by the baseline model $P_b$. Furthermore, according to our empirical results, it will achieve better performance by replacing the sampling operation in Equation~\eqref{eq:sample} with the top-$K$ operation over the distribution $P_b(w|\boldsymbol{x}, \boldsymbol{c})$.

\paragraph{Efficient Inference via Reranking}

As described before, due to the definition of the energy function $S(w, \boldsymbol{x}, \boldsymbol{c})$, it is too costly to evaluate $S(w, \boldsymbol{x}, \boldsymbol{c})$ for all $w$. Thus,
it is infeasible to exactly predict the best $w$ such that $S(w, \boldsymbol{x}, \boldsymbol{c})$ is maximal.
Similar to the top-$K$ operation in the training stage, we adopt it in the inference stage as an approximation.
Specifically, the inference process by the energy-based model includes the following two steps:
\begin{itemize}
    \item Obtain the top-$K$ subset denoted by $\Omega(\boldsymbol{s}, K)$ according to $P_b(w|\boldsymbol{x}, \boldsymbol{c})$, where each element also satisfies the constraint $\boldsymbol{s}$:
    $$\Omega(\boldsymbol{s}, K)=\mathrm{TOP}^K_{w\in \mathcal{V}(\boldsymbol{s})}P_b(w|\boldsymbol{x}, \boldsymbol{c})$$
    \item Output the target word $\hat{w}$ in terms of the energy function as follows:
        \begin{align}
      \hat{w} = \argmax_{w\in \Omega(\boldsymbol{s}, K)} S(w, \boldsymbol{x}, \boldsymbol{c})
        \end{align}
\end{itemize}

\subsection{Weight Initialization via Pre-training}

Recently, pre-trained language models have made exceptional success in numerous natural language processing tasks~\citep{DBLP:conf/naacl/DevlinCLT19,DBLP:conf/acl/LewisLGGMLSZ20,DBLP:conf/nips/Ouyang0JAWMZASR22}.
One of their advantages is that they can learn general and contextual representations to boost the downstream tasks~\citep{DBLP:conf/acl/LiLZWL22,li-autoconv,Shisg}.
Inspired by this, we propose to use our limited supervised bilingual data to conduct a small-scale pre-training for the energy-based model to yield better weight initialization.

Specifically, following practices of Non-Autoregressive Translation \citep{DBLP:conf/emnlp/GhazvininejadLL19,DBLP:conf/acl/LiLZWL22}, we adopt Conditional Masked Bilingual Language Modeling (CMBLM) as our pre-training task.
This CMBLM pre-trained model is supposed to capture bidirectional contextual information better.
Given a sentence pair $(\boldsymbol{x}, \boldsymbol{y})$, similar to masked language models \citep{DBLP:conf/naacl/DevlinCLT19}, we train the model to predict a set of masked target tokens $\boldsymbol{y}_{m}$ given a source sentence $\boldsymbol{x}$ and the observable target words $\boldsymbol{y}_{o} = \boldsymbol{y} \ \backslash \ \boldsymbol{y}_{m}$. The prediction probability distribution for each masked target word $y_i \in \boldsymbol{y}_{m}$ can be formalized as:
\mathchardef\mhyphen="2D
\begin{equation}
    P(y_{i}|\boldsymbol{x},\boldsymbol{y}_{o}) = \mathrm{CMBLM \mhyphen Transformer}(\boldsymbol{x},\boldsymbol{y}_{o})
\end{equation}
As for the model architecture, we adopt the same architecture as $P_b$. 
During the pre-training stage, we randomly mask 15\% of the tokens in $\boldsymbol{y}$ to get $\boldsymbol{y}_m$.
After pre-training, we use the CMBLM pre-trained parameters to initialize our energy-based model.
\section{Experiments}
In this section, we first describe the experimental setup.
Then we report the main results and analyze the proposed approach.

\subsection{Experimental Setup}
\label{experimental_setup}
\paragraph{Datasets}
We experiment on four language pairs: Zh$\Rightarrow$En, En$\Rightarrow$Zh, De$\Rightarrow$En and En$\Rightarrow$De.
For training on Zh$\Rightarrow$En and En$\Rightarrow$Zh, we use training set from the LDC corpus\footnote{The total training set is composed of LDC2002E18, LDC2003E07, LDC2003E14, part of LDC2004T07-08 and LDC2005T06 from \url{https://www.ldc.upenn.edu}\looseness=-1}, which consists of 1.25M sentence pairs.
For training on De$\Rightarrow$En and En$\Rightarrow$De, we use the preprocessed WMT14 dataset by Stanford\footnote{\url{https://nlp.stanford.edu/projects/nmt}}, which consists of 4.5M sentence pairs.
We use the standard validation and test sets released by \citet{DBLP:conf/acl/LiLHS20}\footnote{\url{https://github.com/ghrua/gwlan}}.
Specifically, for Zh$\Rightarrow$En and En$\Rightarrow$Zh, they construct validation set from NIST02 and test set from NIST05 and NIST06.
For De$\Rightarrow$En and En$\Rightarrow$De, they extract validation set from newstest13 and test set from newstest14.

\begin{figure}
    \centering
    \includegraphics[width=\columnwidth]{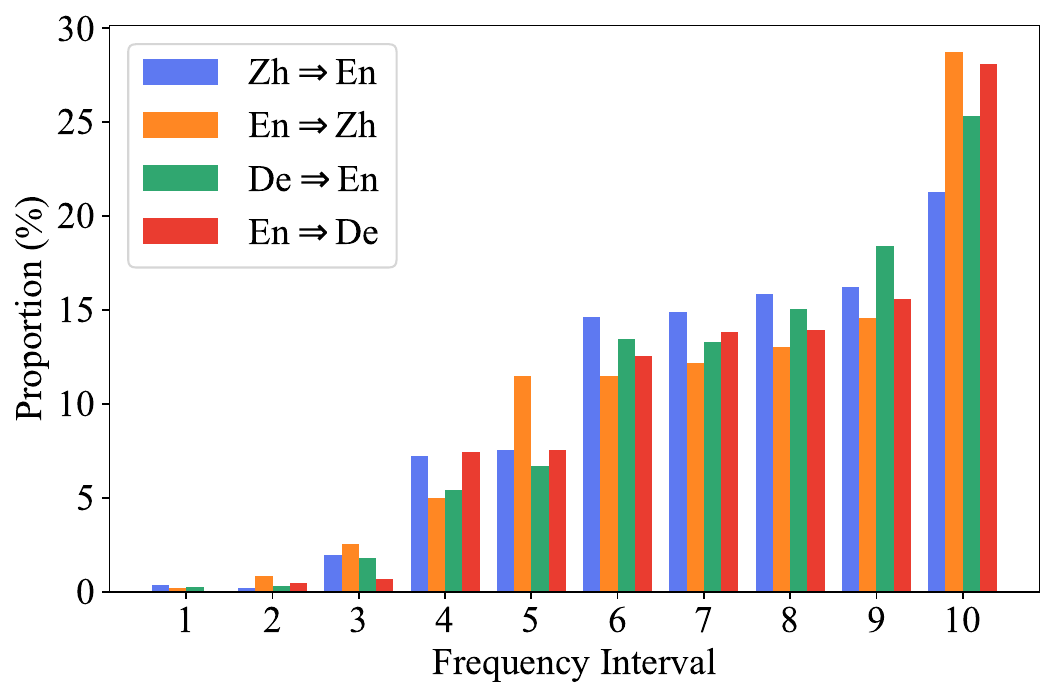}
    \caption{The proportion of different frequency intervals on Zh$\Leftrightarrow$En and De$\Leftrightarrow$En validation datasets. Interval 1 and Interval 10 denote the most frequent interval and the most infrequent interval, respectively.}
    \label{fig:freq_bin}
\end{figure}

In order to construct simulated training data, we follow the same strategy as  \citet{DBLP:conf/acl/LiLHS20} to sample target words, human typed characters and translation context, which aims at avoiding sampling trivial instances.
Statistics of the average length of target words and human typed characters on validation sets are shown in Table~\ref{tab:word_stat}.
As we can see, in general, target words are long and human typed characters are short, which poses a challenge for the WLAC task.
In addition, we also conduct a frequency analysis of each word in training set across four language pairs.
Following this, words are categorized into ten intervals based on their frequency. 
Finally, we calculate the proportion of target words in validation sets corresponding to each frequency interval.
The result is presented in Figure~\ref{fig:freq_bin}.
Figure~\ref{fig:freq_bin} indicates a non-uniform distribution of target words across different frequency intervals.
This data composition basically reflects demands encountered in real-world scenarios,
where non-high frequency words are more challenging for WLAC models.

\begin{table}[t]
\setlength{\tabcolsep}{2.0pt}
\centering
\begin{tabular}{l rrrr}
\toprule
 & \bf{Zh$\Rightarrow$En} & \bf{En$\Rightarrow$Zh}
 & \textbf{De$\Rightarrow$En} & \textbf{En$\Rightarrow$De} \\
\midrule
T.W. & 6.42 & 2.22 & 6.22 & 7.19 \\ 
H.T.C. & 2.00 & 2.05 & 1.95 & 2.20\\ 
\bottomrule
\end{tabular}
\caption{
Statistics of average length of target words and human typed characters on Zh$\Leftrightarrow$En and De$\Leftrightarrow$En validation sets. T.W. and H.T.C are short for target words and human typed characters, respectively.
% ‘$^*$’: H.T.C for Chinese is pinyin.
}
\label{tab:word_stat}
\end{table}

\paragraph{Baselines}
We compare our model with the following baseline models:
\begin{itemize}[itemsep=0.1pt]
    \item \textsc{TransTable}: A statistical method inspired by \citet{DBLP:conf/ijcai/HuangZZZ15}. They create a word-level translation table with a word alignment toolkit\footnote{\url{https://github.com/clab/fast_align}}. During the inference stage, they use the translation table to get translations of all source words and filter out invalid candidate words through human typed characters. Ultimately, they pick the candidate word with the highest frequency as the prediction.
    
    \item \textsc{Trans-PE}: A Transformer-based baseline inspired by \citet{langlais2000transtype};\citet{DBLP:conf/emnlp/SantyDCB19}. They first train a vanilla Transformer on training set. While testing, they only feed the left translation context to the Transformer decoder. Then they conduct a next-word prediction task with human typed characters as hard constraints to get the prediction word.
    
    \item \textsc{Trans-NPE}: The only difference between this method between \textsc{Trans-PE} is that there is no position encoding layer in the decoder of \textsc{Trans-NPE}. They apply average pooling to the representations of all translation context words. And then, they use the pooled representation to predict the target word.
    
    \item \textsc{\pb}: The word prediction model defined in Equation~\eqref{eq:disc}, which is the state-of-the-art model of the WLAC task.\looseness=-1

    \item \textsc{Trans-BPE}: Inspired by~\citet{DBLP:conf/iclr/CaoI0P21,DBLP:conf/emnlp/YangMZL022}, we also implement a new Transformer-based baseline over subwords.
    Specifically, we apply BPE to segment words into subwords.
    During the inference stage, we adopt Prefix-Constrained Beam Search \citep{DBLP:conf/iclr/CaoI0P21} to generate outputs which start with human typed characters.
    This model is expected to be capable of defining the hidden vector on top of previously generated subwords and the input context to predict the next subword.\looseness=-1

\end{itemize}

% On the Language Coverage Bias for Neural Machine Translation
\begin{table*}[t]
\centering
\begin{tabular}{ll cc cc cc cc}
\toprule
\multirow{2}{*}{\bf \#} & \multirow{2}{*}{\bf Systems}  &   \multicolumn{2}{c}{\bf Zh$\Rightarrow$En}  & \multicolumn{2}{c}{\bf En$\Rightarrow$Zh} & \multicolumn{2}{c}{\bf De$\Rightarrow$En} & \multicolumn{2}{c}{\bf En$\Rightarrow$De}\\
\cmidrule(lr){3-4}\cmidrule(lr){5-6}\cmidrule(lr){7-8}\cmidrule(lr){9-10}
& & \multicolumn{1}{c}{\bf NIST05}  & \multicolumn{1}{c}{\bf NIST06} & \multicolumn{1}{c}{\bf NIST05}  & \multicolumn{1}{c}{\bf NIST06} & \multicolumn{1}{c}{\bf NT13}  & \multicolumn{1}{c}{\bf NT14} & \multicolumn{1}{c}{\bf NT13}  & \multicolumn{1}{c}{\bf NT14} \\
\midrule
% \multicolumn{10}{l}{\it{Reported results in previous work}} \\
% \midrule
1 & \textsc{TransTable$^\dagger$} & 41.40                      & 39.78                      & 28.00                      & 26.99                      &     37.43        &       36.64     &       32.99      &    31.12        \\
2 & \textsc{Trans-PE$^\dagger$}   & 34.51                      & 35.50                      & 32.23                      & 34.88                      &        34.45     &      33.02      &      31.51      &          30.65  \\
3 & \textsc{Trans-NPE$^\dagger$}  & 35.97                      & 36.78                      & 34.31                      & 36.19                      &         36.69    &      36.01      &       33.25      &        31.30    \\
4 & \textsc{\pb$^\dagger$} & 55.54                      & 55.85                      & 53.64                     & 54.25                             &       57.84     &    56.75   &    \underline{56.91}     &      52.68      \\
5 & \textsc{\pb$^*$} & 55.52 & 56.57 & \underline{53.89} & 54.24 & 59.11 & 56.99 & 56.89 & 53.80 \\
6 & \textsc{Trans-BPE$^*$} & \underline{57.29} & \underline{57.80} & 53.82 & \underline{55.93} & \underline{61.44} & \underline{59.95} & 55.41 & \underline{54.80} \\
7 & \textsc{Ours$^*$} & \textbf{65.61} & \textbf{65.44} & \textbf{60.43} & \textbf{61.25} & \textbf{64.62} & \textbf{63.13} & \textbf{62.23} & \textbf{60.24} \\ \bottomrule
\end{tabular}
\caption{The main results of different systems on Zh$\Leftrightarrow$En and De$\Leftrightarrow$En datasets. The results in this table are the average accuracy across four translation context types (i.e., zero-context, prefix, suffix and bi-context). ‘$^\dagger$’: results are reported in previous work. ‘$^*$’: results are implemented by ourselves, which is the average of 5 runs with different random seeds. The best and the second-best results are in \textbf{bold} and \underline{underlined} fonts, respectively.}
\label{tab:main}
\end{table*}
%-----------------------------------------------

\begin{table*}[t]
  \resizebox{2.08\columnwidth}{!}{
\begin{tabular}{ll ccccc ccccc}
\toprule
  \multirow{2}{*}{\textbf{\#}} &    \multirow{2}{*}{\textbf{Systems}}         & \multicolumn{5}{c}{\textbf{Zh$\Rightarrow$En}} & \multicolumn{5}{c}{\textbf{En$\Rightarrow$Zh}} \\ \cmidrule(lr){3-7}\cmidrule(lr){8-12}
 &            & \multicolumn{1}{l}{\textbf{Prefix}} & \multicolumn{1}{c}{\textbf{Suffix}} & \multicolumn{1}{c}{\textbf{Zero.}} & \multicolumn{1}{c}{\textbf{Bi.}} & \multicolumn{1}{c}{\textbf{Overall}} & \multicolumn{1}{c}{\textbf{Prefix}} & \multicolumn{1}{c}{\textbf{Suffix}} & \multicolumn{1}{c}{\textbf{Zero.}} & \multicolumn{1}{c}{\textbf{Bi.}} & \multicolumn{1}{c}{\textbf{Overall}} \\\midrule
1  & \textsc{TransTable$^\dagger$} & 41.91                      & 44.99                      & 44.19                            & 43.28                          & 43.59                   & 29.73                      & 32.80                       & 29.73                            & 29.61                          & 30.46                   \\
2  & \textsc{Trans-PE$^\dagger$}   & 29.84                      & 38.61                      & 26.08                            & 48.06                          & 35.64                   & 30.64                      & 34.97                      & 22.67                            & 38.95                          & 31.80                   \\
3  & \textsc{Trans-NPE$^\dagger$}  & 37.36                      & 40.43                      & 29.50                             & 44.42                          & 37.92                   & 36.10                       & 43.05                      & 32.00                               & 45.79                          & 39.23                   \\
    % 4  & \textsc{\pb-Sep}    & 58.43 & 60.59 & 53.99 & 64.46 & 59.36 & 60.02 & 61.05 & 53.76 & 64.46 & 59.82 \\
    4  & \textsc{\pb$^\dagger$}  & 59.91 & 60.71 & \underline{55.35} & 62.30 & 59.56 & 61.39 & 61.73 & 53.87 & 63.78 & 60.19 \\
    5 & \textsc{\pb$^*$} & 58.59 & 63.34 & 54.35 & 68.21 & 61.12 & 60.47 & \underline{62.94} & 53.40 & 67.40 & 61.05 \\
    6 & \textsc{Trans-BPE$^*$} & \underline{60.14} & \underline{64.03} & 55.24 & \underline{69.84} & \underline{62.31} & \underline{61.89} & 62.54 & \underline{55.02} & \underline{69.26} & \underline{62.18} \\
    7 & \textsc{Ours$^*$} & \textbf{68.13} & \textbf{70.32} & \textbf{66.45} & \textbf{75.56} & \textbf{70.12} & \textbf{68.63} & \textbf{69.16} & \textbf{59.91} & \textbf{71.80} & \textbf{67.37} \\ \bottomrule
\end{tabular}
}
\caption{The detailed results for each translation context type of different systems on Zh$\Leftrightarrow$En validation set.
% ‘$^\dagger$’: results are reported in previous work. ‘$^*$’: results are implemented by ourselves. The best and the second-best results are in \textbf{bold} and \underline{underlined} fonts, respectively.
}
\label{tab:nist02}
\end{table*}

%-----------------------------------------------

\begin{table*}[t]
  \resizebox{2.08\columnwidth}{!}{
\begin{tabular}{ll ccccc ccccc}
\toprule
  \multirow{2}{*}{\textbf{\#}} &    \multirow{2}{*}{\textbf{Systems}}         & \multicolumn{5}{c}{\textbf{De$\Rightarrow$En}} & \multicolumn{5}{c}{\textbf{En$\Rightarrow$De}} \\ \cmidrule(lr){3-7}\cmidrule(lr){8-12}
 &            & \multicolumn{1}{l}{\textbf{Prefix}} & \multicolumn{1}{c}{\textbf{Suffix}} & \multicolumn{1}{c}{\textbf{Zero.}} & \multicolumn{1}{c}{\textbf{Bi.}} & \multicolumn{1}{c}{\textbf{Overall}} & \multicolumn{1}{c}{\textbf{Prefix}} & \multicolumn{1}{c}{\textbf{Suffix}} & \multicolumn{1}{c}{\textbf{Zero.}} & \multicolumn{1}{c}{\textbf{Bi.}} & \multicolumn{1}{c}{\textbf{Overall}} \\\midrule
    1 & \textsc{\pb} & 57.52 & 61.59 & \underline{51.01} & 66.32 & 59.11 & \underline{54.63} & 60.83 & \underline{48.51} & \underline{63.58} & \underline{56.89} \\
    2 & \textsc{Trans-BPE} & \textbf{61.88} & \underline{65.35} & 50.68 & \underline{67.84} & \underline{61.44} & 52.25 & \underline{60.94} & 46.60 & 61.85 & 55.41 \\
    3 & \textsc{Ours} & \underline{61.47} & \textbf{68.01} & \textbf{58.47} & \textbf{70.54} & \textbf{64.62} & \textbf{57.17} & \textbf{67.01} & \textbf{56.45} & \textbf{68.28} & \textbf{62.23 }\\ \bottomrule
\end{tabular}
}
\caption{The detailed results for each translation context type of different systems on De$\Leftrightarrow$En validation set.}
% \caption{The detailed results of different systems on De$\Leftrightarrow$En validation set. ‘$^*$’: results are implemented by ourselves. The best and the second-best results are in \textbf{bold} and \underline{underlined} fonts, respectively.}
\label{tab:nt13}
\end{table*}

\paragraph{Implementation Details}
We implement our energy-based model on top of the Transformer-Base architecture \citep{DBLP:conf/nips/VaswaniSPUJGKP17} implemented in \texttt{Fairseq} toolkit ~\citep{DBLP:conf/naacl/OttEBFGNGA19}\footnote{\url{https://github.com/facebookresearch/fairseq}}. The source encoder is a stack of 6 Transformer encoder blocks. The target encoder is also composed of 6 blocks, each of which is a Transformer encoder block with an additional cross-attention layer between the multi-head self-attention layer and feed-forward layer.
The vocabulary size is 60K for Chinese, 50K for German and 50K for English.
As for the implementation of \textsc{Trans-BPE}, we adopt the Transformer-Base architecture and make adjustments to the input of Transformer Encoder. 
Specifically, we feed the concatenation of the source context, target context and placeholder \texttt{[MASK]} to the Transformer Encoder, and adopt segment embedding to distinguish different languages as \citet{DBLP:conf/emnlp/YangMZL022}.
The vocabulary size is 32K for both Zh$\Leftrightarrow$En and De$\Leftrightarrow$En.
For a fair comparison, we also re-implement \pb~with the same hyperparameter settings as the energy-based model.\looseness=-1

For above models,
we set $d_{model}$ = 512, $d_{hidden}$ = 2048, $n_{head}$ = 8 and $p_{dropout}$ = 0.1.
And the learning rate is set as 0.0005, the warmup step is set as 4,000 steps.
All models are trained with 4096 tokens per batch for a maximum of 50,000 steps with Adam optimizer~\citep{DBLP:journals/corr/KingmaB14} on 8 NVIDIA V100 GPUs.
We update the model parameters after accumulating 2 gradients for \textsc{Trans-BPE} and 1 gradient for \pb~and \textsc{Ours}.
Models are selected with the best accuracy on the validation set.
We repeat the main experiment 5 times by using different random seeds.

\subsection{Main Results}

\paragraph{Evaluation on Word Prediction by ACC}
Table~\ref{tab:main} lists the main results on four language pairs.
From the table, we can make three observations:
First, statistical and intuitive Transformer-based methods (\#1-3) perform poorly on all language pairs.
We speculate that this is because these approaches can not make full use of the information from the input context (e.g., source sentence).
Second, \textsc{Trans-BPE} outperforms \textsc{\pb} on average accuracy.
The reason behind this could be attributed to the effectiveness of \textsc{Trans-BPE} to leveraging more valuable source sentence information than \pb, which we will elaborate on in Section \ref{sec:5.4}.
Third, our energy-based model (\#7) improves over the previous SOTA performance by an average of \textbf{\improvement} accuracy points across all language pairs, which demonstrates its effectiveness.
Furthermore, in Table~\ref{tab:nist02} and Table~\ref{tab:nt13}, we report the detailed results of different systems on four translation context types on the Zh$\Leftrightarrow$En and De$\Leftrightarrow$En validation sets.
We can find that, our energy-based model can almost achieve performance improvement on each translation context type, except for De$\Rightarrow$En prefix context, and finally results in overall performance in Table~\ref{tab:main}.\looseness=-1

\paragraph{Human Evaluation}
\label{sec:5.2.1}
% \begin{table*}[t]
%   \resizebox{2.08\columnwidth}{!}{
% \begin{tabular}{ll ccccc ccccc}
% \toprule
%   \multirow{2}{*}{\textbf{\#}} &    \multirow{2}{*}{\textbf{Systems}}         & \multicolumn{5}{c}{\textbf{Zh$\Rightarrow$En}} & \multicolumn{5}{c}{\textbf{En$\Rightarrow$Zh}} \\ \cmidrule(lr){3-7}\cmidrule(lr){8-12}
%  &            & \multicolumn{1}{l}{\textbf{Prefix}} & \multicolumn{1}{c}{\textbf{Suffix}} & \multicolumn{1}{c}{\textbf{Zero.}} & \multicolumn{1}{c}{\textbf{Bi.}} & \multicolumn{1}{c}{\textbf{Overall}} & \multicolumn{1}{c}{\textbf{Prefix}} & \multicolumn{1}{c}{\textbf{Suffix}} & \multicolumn{1}{c}{\textbf{Zero.}} & \multicolumn{1}{c}{\textbf{Bi.}} & \multicolumn{1}{c}{\textbf{Overall}} \\\midrule
%     1 & \textsc{$P_b$} & 80.50 & 82.00 & 87.50 & 84.00 & 83.5000 & 78.50 & 84.50 & 86.00 & 82.50 & 82.8750 \\
%     2 & \textsc{Trans-BPE} & 78.50 & 83.50 & 86.50 & 94.00 & 85.6250 & \textbf{86.50} & 82.50 & 89.00 & 81.00 & 84.7500 \\
%     3 & \textsc{Ours} & \textbf{90.50} & \textbf{86.50} & \textbf{88.00} & \textbf{94.50} & \textbf{89.8750} & \textbf{86.50} & \textbf{86.50} & \textbf{92.50} & \textbf{88.00} & \textbf{88.3750} \\ \bottomrule
% \end{tabular}
% }
% \caption{The detailed results of different systems under the Zh$\Rightarrow$En and En$\Rightarrow$Zh human evaluation setting.}
% \label{tab:human_evaluation}
% \end{table*}

\begin{table*}[t]
  \resizebox{2.08\columnwidth}{!}{
\begin{tabular}{ll ccccc ccccc}
\toprule
  \multirow{2}{*}{\textbf{\#}} &    \multirow{2}{*}{\textbf{Systems}}         & \multicolumn{5}{c}{\textbf{Zh$\Rightarrow$En}} & \multicolumn{5}{c}{\textbf{En$\Rightarrow$Zh}} \\ \cmidrule(lr){3-7}\cmidrule(lr){8-12}
 &            & \multicolumn{1}{l}{\textbf{Prefix}} & \multicolumn{1}{c}{\textbf{Suffix}} & \multicolumn{1}{c}{\textbf{Zero.}} & \multicolumn{1}{c}{\textbf{Bi.}} & \multicolumn{1}{c}{\textbf{Overall}} & \multicolumn{1}{c}{\textbf{Prefix}} & \multicolumn{1}{c}{\textbf{Suffix}} & \multicolumn{1}{c}{\textbf{Zero.}} & \multicolumn{1}{c}{\textbf{Bi.}} & \multicolumn{1}{c}{\textbf{Overall}} \\\midrule
    1 & \textsc{$P_b$} & 81.50 & 82.50 & 87.00 & 83.00 & 83.50 & 79.50 & 84.00 & 86.50 & 83.50 & 83.38 \\
    2 & \textsc{Trans-BPE} & 80.00 & 84.00 & 86.50 & 94.00 & 86.13 & 86.00 & 84.50 & 89.50 & 80.00 & 85.00 \\
    3 & \textsc{Ours} & \textbf{90.50} & \textbf{87.00} & \textbf{88.00} & \textbf{94.50} & \textbf{90.00} & \textbf{86.50} & \textbf{87.00} & \textbf{93.50} & \textbf{88.50} & \textbf{88.88} \\ \bottomrule
\end{tabular}
}
\caption{The detailed results of different systems under the Zh$\Rightarrow$En and En$\Rightarrow$Zh human evaluation setting.
The results in the table represent the average rating scores from two evaluators.}
\label{tab:human_evaluation}
\end{table*}
It is also crucial to assess the actual improvement in effectiveness of our approach via human evaluation.
However, performing comprehensive human evaluations can be resource-intensive in terms of labor.
As a compromise, we randomly sample 400 examples from the original Zh$\Rightarrow$En and En$\Rightarrow$Zh NIST05 test sets, with 100 instances for each translation context type.
We then collect predictions from three models: $P_b$, \textsc{TransBPE} and \textsc{Ours}.
Subsequently, we enlist two professional evaluators to assess the appropriateness of predictions of these models.
The human evaluators are presented with the input context, human typed characters as well as each prediction.
The predictions, originating from different models, are anonymized to the evaluators.
The human evaluators are asked to assign binary scores for each prediction, where a score of `1' indicates appropriateness, while `0' signifies inappropriateness.
Results of human evaluation are presented in Table~\ref{tab:human_evaluation}.
The Cohen's kappa is 0.92 between the two translators, which is a relatively high agreement.
Table~\ref{tab:human_evaluation} demonstrates that our energy-based model retains an advantage over previous methods under human evaluation.
What's more, one detail worth noting is that, compared to results in Table~\ref{tab:main}, all models exhibit an improvement in performance when evaluated manually.
This can be attributed to the fact that the accuracy metric only considers the top-1 prediction, while other predictions may also be valid.
To ensure consistency with prior research, we utilize accuracy as the evaluation metric in the following sections.

% opus100
% \begin{table*}[t]
% \setlength{\tabcolsep}{3pt}
% \centering
% \begin{tabular}{l rr rr}
% \toprule
% \multirow{2}{*}{\bf Model}  
% %& \multicolumn{4}{c}{\bf OPUS-100 Data}\\
% %\cmidrule(lr){2-5}
% & \multicolumn{2}{c}{\bf Zh$\Rightarrow$En}  & \multicolumn{2}{c}{\bf De$\Rightarrow$En}\\
% \cmidrule(lr){2-3}\cmidrule(lr){4-5}
%     &   \bf NIST05  & \bf NIST06    &   \bf NT15  & \bf NT16 \\
% \midrule
% \multicolumn{5}{c}{{ \bf \em \textsc{Ours}}}\\
% Raw              &  65.39   & 65.47     &   64.76 & 63.18       \\
% \hdashline
% \quad w/o Reranker         &  55.55   &  56.27   &  59.03 & 57.24      \\
% %\quad w/o CMLM Pretraining &   50     &   50     &   50     \\
% \quad w/ Generator Init    &   64.43     &   58.09    &   60.15 & 58.03      \\
% \midrule
% \multicolumn{5}{c}{{ \bf \em \textsc{WPM-SEP}}}\\
% Raw           &      55.55     &   56.27  &  59.03 & 57.24   \\
% \hdashline
% \quad w/ CMLM Pretraining       &     59.45     &    60.67 & 60.15 & 59.33   \\
% \bottomrule
% \end{tabular}
% \caption{
% {
% Results of MNMT models trained on the OPUS-100 massively multilingual translation dataset.}
% }
% \label{tab:main-bleu-otr-opus100}
% \end{table*}

\begin{table*}[t]
\centering
\begin{tabular}{l cc cc cc cc}
\toprule
\multirow{3}{*}{\bf Systems}  &   \multicolumn{4}{c}{\bf Zh$\Rightarrow$En}  & \multicolumn{4}{c}{\bf De$\Rightarrow$En}\\
\cmidrule(lr){2-5}\cmidrule(lr){6-9}
& \multicolumn{2}{c}{\bf NIST05}  & \multicolumn{2}{c}{\bf NIST06} &  \multicolumn{2}{c}{\bf NT13}  & \multicolumn{2}{c}{\bf NT14}\\
\cmidrule(lr){2-3}\cmidrule(lr){4-5}\cmidrule(lr){6-7}\cmidrule(lr){8-9}
    &   \bf Acc.  & \bf $\triangle$    &   \bf Acc.  & \bf $\triangle$    &   \bf Acc.  & \bf $\triangle$    &   \bf Acc.  & \bf $\triangle$ \\
\midrule
% \multicolumn{9}{c}{{ \bf \em \textsc{Ours}}}\\

\textbf{\textsc{$\boldsymbol{P_b}$}}           &      55.52  & -  &   56.57  & - &  59.11 & - & 56.99  & - \\
% \hdashline
\quad w/ CMBLM       &     59.45 &  +3.93  &    60.67 & +4.10 & 60.83 & +1.72  & 59.33  & +2.34 \\
\midrule
\textbf{\textsc{Ours} } w/ \pb~Init    &   58.09  &  +2.57 &   58.54 & +1.97  &   60.15 & +1.04 & 58.03   & +1.04   \\
% \hdashline
 \quad  w/ CMBLM                &  65.61  & +10.09  & 65.44  &  +8.87  &   64.62 & +5.51 & 63.13 & +6.14      \\

%\quad w/o Reranker         &  55.55 & -9.84  &  56.27  & -9.20 &  59.03 & -5.73 & 57.24 & -5.94     \\

% \multicolumn{9}{c}{{ \bf \em \textsc{WPM-Sep}}}\\

\bottomrule
\end{tabular}
\caption{Performance of weight initialization on Zh$\Rightarrow$En and De$\Rightarrow$En datasets. The results in this
table are the average accuracy across four translation context types.}
\label{tab:ablation}
\end{table*}
\subsection{Ablation Studies}
\paragraph{Negative Sampling for Training}
% \begin{table}[t]
% \setlength{\tabcolsep}{3pt}
% \centering
% \begin{tabular}{l rrrr}
% \toprule
% \bf{Strategy} & \bf{NIST02} & \bf{NIST05} & \bf{NIST06} & \textbf{P}$\boldsymbol{_b}$ \\
% \midrule
% Uniform & 66.71 & 62.22 &  62.92 & - \\
% \midrule
% Random & 69.10 & 64.97 & 64.47 & \textsc{WPM-Sep} \\
% Top-$p$ & 69.55 &  64.84 &  64.97 & \textsc{WPM-Sep} \\
% Top-$K$ & \bf 70.62 & \bf 65.39 & \bf 65.47 & \textsc{WPM-Sep} \\
% \bottomrule
% \end{tabular}
% \caption{
% The results of different negative sampling techniques on Zh$\Rightarrow$En. The results in this
% table are the averaged accuracy on four translation context types (i.e., zero-context, prefix, suffix and bi-context).
% }
% \label{tab:negative_sampling_methods}
% \end{table}

\begin{table}[t]
\setlength{\tabcolsep}{3pt}
\centering
\begin{tabular}{ll rrr}
\toprule
\textbf{Dist.} &\bf{Strategy} & \bf{NIST02} & \bf{NIST05} & \bf{NIST06}  \\
\midrule
Uniform & Random & 66.71 & 62.22 &  62.92 \\
\midrule
\multirow{3}{*}{\pb}& Random & 69.10 & 64.97 & 64.47 \\
& Top-$p$ & 69.55 &  64.84 &  64.97 \\
& Top-$K$ & \bf 70.12 & \bf 65.61 & \bf 65.44 \\
\bottomrule
\end{tabular}
\caption{
The results of different negative sampling strategies on Zh$\Rightarrow$En. The results in this table are the average accuracy across four translation context types.
}
\label{tab:negative_sampling_methods}
\end{table}
As we state in Section~\ref{sec:3}, negative sampling in the training stage can affect the performance of the energy-based model.
We consider two sampling distributions (the uniform distribution and the distribution of \pb ) and three negative sampling strategies, i.e., random sampling, top-$p$ sampling and top-$K$ sampling.
We compare them on Zh$\Rightarrow$En dataset.
During the inference stage, we use $P_b$ to recall top-$8$ predicted words as candidate target words for these models trained with different negative sampling techniques.

We report the results in Table~\ref{tab:negative_sampling_methods}.
We can observe that the random sampling strategy from the uniform distribution is not as effective as the other three sampling configurations from $P_b$. We conjecture that negative samples by random sampling on the uniform distribution could be too trivial to recognize hard negatives, which may hinder the performance of the energy-based model. While sampling according to $P_b$ (i.e., the other three strategies) can sample hard negatives and facilitate the training of the energy-based model.

\begin{figure}
    \centering
    \includegraphics[width=\columnwidth]{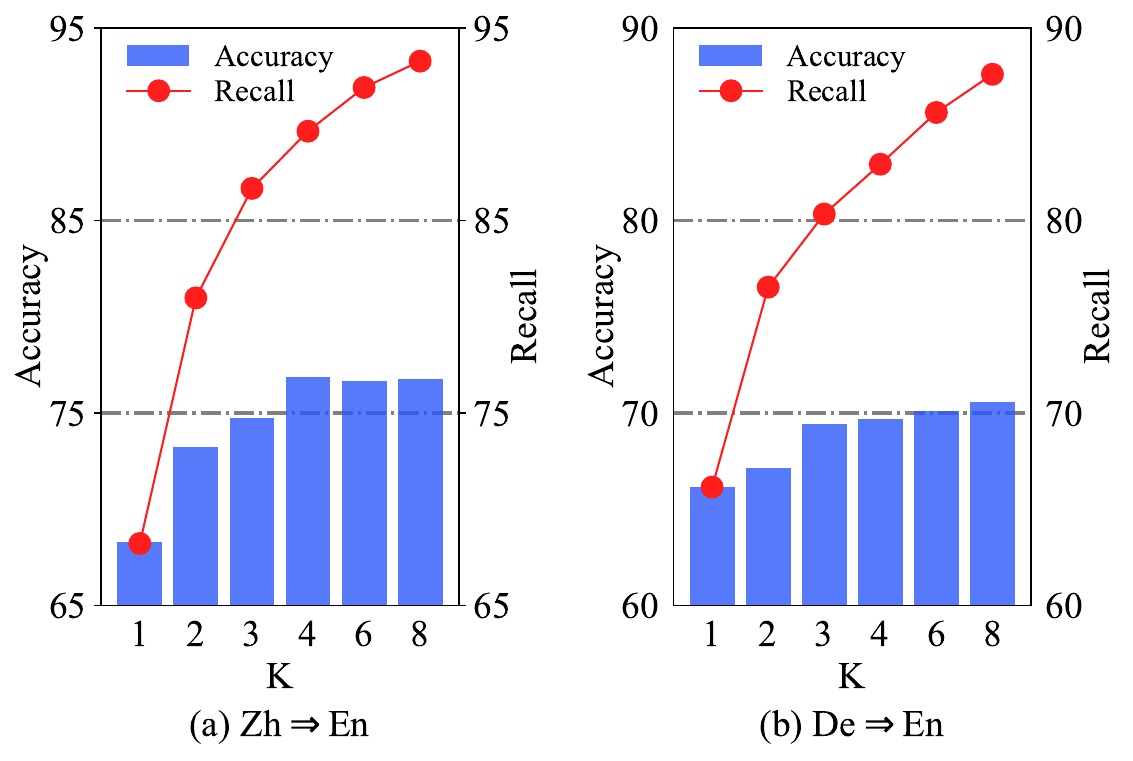}
    \caption{Accuracy of our energy-based model and recall of ground-truth word with different $K$ on Zh$\Rightarrow$En NIST02 dataset (a) and De$\Rightarrow$En NT13 dataset (b). Experiments are conducted in the bi-context scenario.}
    \label{fig:vary_k_acc_recall}
\end{figure}

\paragraph{$K$-best Size in Inference}
We further analyze the impact of candidate word set size $K=\mathcal{V}(\boldsymbol{s})$ during the inference with the energy-based model.
Figure~\ref{fig:vary_k_acc_recall} shows that, as $K$ increases, the accuracy improvement increases rapidly from $K=1$ to $K=4$ and starts to saturate after $K=4$.
The recall of the ground-truth word shares the same trend as accuracy: it first improves sharply, then increases slowly and reaches a relatively high value. So for the efficiency and effectiveness trade-off,  we choose to use $K=8$ as our candidate word set size in all experiments during the inference.\looseness=-1

\paragraph{Weight Initialization}
Our energy-based model is pre-trained by a CMBLM pre-training strategy.
Therefore, its improvements might come from two aspects, including 1) the energy-based model and 2) better initialization weights and representations learned from the CMBLM pre-training task.
Hence, we perform further studies to quantify the contribution of each component of our approach. To this end, we conduct two experiments: we replace the CMBLM pre-training by initializing the weights from the baseline WPM $P_b$; and we apply the CMBLM pre-training on top of $P_b$ and compare it with the energy-based model with the CMBLM pre-training.
We evaluate all these methods on Zh$\Rightarrow$En dataset and De$\Rightarrow$En dataset and present the results in Table~\ref{tab:ablation}.\looseness=-1

The results in Table~\ref{tab:ablation} illustrate that: First, initializing the weights of the energy-based model with $P_b$ is not as effective as initializing with the CMBLM pre-training strategy. Second, although both $P_b$ and our energy-based model benefit from the CMBLM pre-training strategy, the gain for the energy-based model is much larger. These observations demonstrate that a simple pre-training method can not activate the potential of the energy-based model and the CMBLM pre-training strategy succeeds.\looseness=-1

\subsection{Analysis}
\label{sec:5.4}
\paragraph{Evaluation on Prefix-Decoding and Post-Editing Settings}
% On the Language Coverage Bias for Neural Machine Translation
\begin{table*}[t]
\centering
\begin{tabular}{ll cc cc cc cc}
\toprule
\multirow{2}{*}{\bf \#} & \multirow{2}{*}{\bf Systems}  &   \multicolumn{2}{c}{\bf Zh$\Rightarrow$En}  & \multicolumn{2}{c}{\bf En$\Rightarrow$Zh} & \multicolumn{2}{c}{\bf De$\Rightarrow$En} & \multicolumn{2}{c}{\bf En$\Rightarrow$De}\\
\cmidrule(lr){3-4}\cmidrule(lr){5-6}\cmidrule(lr){7-8}\cmidrule(lr){9-10}
& & \multicolumn{1}{c}{\bf NIST05}  & \multicolumn{1}{c}{\bf NIST06} & \multicolumn{1}{c}{\bf NIST05}  & \multicolumn{1}{c}{\bf NIST06} & \multicolumn{1}{c}{\bf NT13}  & \multicolumn{1}{c}{\bf NT14} & \multicolumn{1}{c}{\bf NT13}  & \multicolumn{1}{c}{\bf NT14} \\
\midrule
\multicolumn{10}{c}{\it{Prefix-Decoding}} \\
\midrule
1 & \textsc{\pb} & 79.57                     & \underline{78.85}                    & 73.45                     & 74.95                             &       81.41     &    79.15   &    76.09     &     73.38      \\
2 & \textsc{Trans-BPE} & \underline{80.96} & 78.63 & \underline{74.47} & \underline{75.28} & \underline{81.99} & \underline{79.63} & \underline{77.66} & \underline{74.23}\\
3 & \textsc{Ours} & \textbf{83.73} & \textbf{83.21} & \textbf{77.34} & \textbf{79.10} & \textbf{84.13} & \textbf{82.60} & \textbf{78.68} & \textbf{76.73} \\
\midrule
\multicolumn{10}{c}{\it{Post-Editing}} \\
\midrule
1 & \textsc{\pb} & 85.30                      & 86.95                     & 80.11                    & \underline{80.93}                             &       86.79     &    83.70   &    83.86     &     79.82      \\
2 & \textsc{Trans-BPE} & \underline{85.95} & \underline{87.53} & \underline{81.96} & 80.73 & \underline{87.81} & \underline{84.84} & \underline{85.01} & \underline{80.93}\\
3 & \textsc{Ours} & \textbf{89.74} & \textbf{90.16} & \textbf{84.09} & \textbf{84.16} & \textbf{89.85} & \textbf{87.04} & \textbf{86.99} & \textbf{83.02} \\
\bottomrule
\end{tabular}
\caption{The main results of different systems on Zh$\Leftrightarrow$En and De$\Leftrightarrow$En datasets under prefix-decoding and post-editing settings.}
\label{tab:imt_ts}
\end{table*}
Although our work mainly focuses on four translation context types in the WLAC task, we also explore whether the energy-based model would still improve performance on two common translation scenarios including prefix-decoding widely used in left-to-right interactive machine translation and post-editing as stated in Section~\ref{sec:2.1}.
To this end, we implement \pb, \textsc{Trans-BPE} and \textsc{Ours} on these two scenarios with the same parameter configuration in Section~\ref{experimental_setup}.
As for the construction of validation sets and test sets,
we adopt the same simulation method as \citet{DBLP:conf/acl/LiLHS20} other than that the target word must be consecutive to target context.
Table~\ref{tab:imt_ts} shows the results of \pb , \textsc{Trans-BPE} and \textsc{Ours} on prefix-decoding and post-editing scenarios.
As we can see, \textsc{Ours} can further improve average accuracy points across all language pairs by 3.22 on post-decoding and by 2.68 on post-editing,
demonstrating the effectiveness of our energy-based model.

\paragraph{Evaluation on Usage of Informative Context}
\label{usage_of_informative_context}
\begin{figure}
    \centering
    \includegraphics[width=\columnwidth]{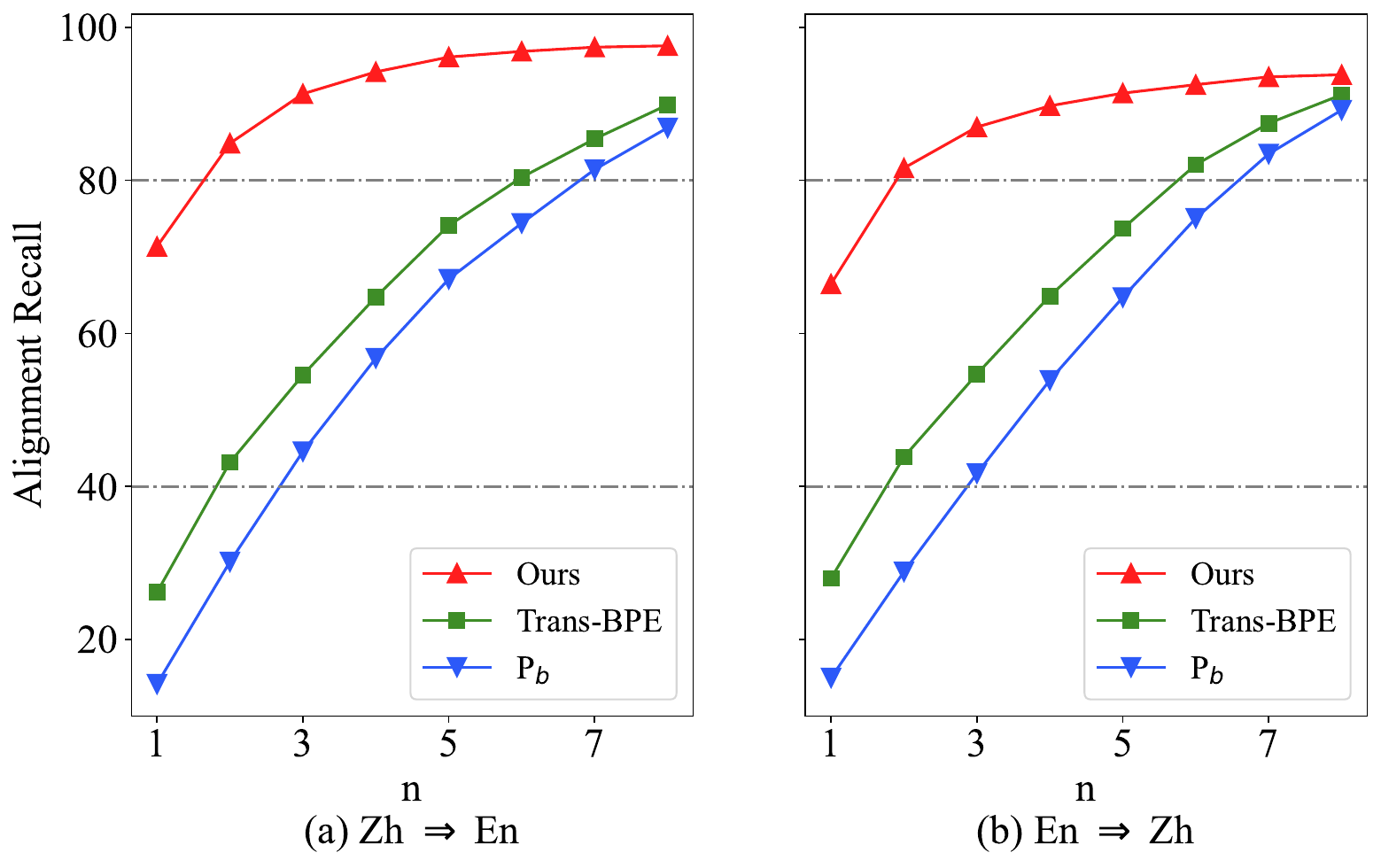}
    \caption{Alignment recall@$n$ on Zh$\Leftrightarrow$En NIST05 dataset with $n$ ranging from $1$ to $8$. Experiments are conducted in the bi-context scenario.
    }
    \label{fig:word_alignment}
\end{figure}
As we have claimed in Section~\ref{sec:3}, our motivation is that the energy-based model is capable of capturing more informative context for word prediction, which thereby leads to better performance eventually.
In addition to the intuitive example in Figure~\ref{fig:intro}(c), we design an automatic metric to verify our motivation. This metric is inspired by the word alignment error rate for the cross-attention in the Transformer~\citep{DBLP:conf/acl/LiLLMS19,DBLP:conf/emnlp/GargPNP19}. 
Specifically, as shown in Figure~\ref{fig:intro}(c), the metric (alignment recall@$n$) is defined as the recall rate of the informative source word ``\emph{Krankheit}'' by the top-$n$ source words according to the attention score by the Transformer architecture. 
For each ground-truth target word, e.g., ``\emph{disease}'' in Figure~\ref{fig:intro}(c), the informative source word is defined by the manually annotated word alignment.

We use the human-annotated alignment data on Zh$\Leftrightarrow$En NIST05 dataset and conduct experiments in the bi-context scenario. 
We compare the alignment recall@$n$ between \textsc{\pb}, \textsc{Trans-BPE} and \textsc{Ours} in Figure~\ref{fig:word_alignment}. 
As we can see, the alignment recall@$1$ of \textsc{Ours} is higher than \textsc{\pb} by 60 points and when $n$ is small, it always maintains this advantage.
What's more, \textsc{Trans-BPE} also achieves better alignment recall@$n$ than~\pb.
This may serve as quantitative evidence that introducing subwords or the entire candidate target word into the modeling of hidden vectors with the input context, as implemented in \textsc{Trans-BPE} and \textsc{Ours}, can make more use of informative context than \textsc{\pb}~\citep{DBLP:conf/iclr/CaoI0P21}.
And results illustrated in the Figure~\ref{fig:word_alignment} also reveal that our energy-based model might be more effective in leveraging informative context than \textsc{Trans-BPE}.

\paragraph{Error Analysis}
\begin{table}[t]
\setlength{\tabcolsep}{4pt}
\centering
\begin{tabular}{l cccc}
\toprule
\bf{Systems} &\bf{Type-I} & \bf{Type-II} & \bf{Type-III} & \bf{Total}  \\
\midrule
\pb & 79 & 29 & 20 & 128 \\
\midrule
\textsc{Ours} & 57~\scriptsize{(-25)} & 11~\scriptsize{(-20)} & 9~\scriptsize{(-14)} & 77  \\
% \textsc{Ours} & 57~\scriptsize{(54+3)} & 11~\scriptsize{(9+2)} & 9~\scriptsize{(6+3)} & 77  \\
\bottomrule
\end{tabular}
\caption{
Quantitative results of error occurrences between \pb~and \textsc{Ours}.
% The numbers in parentheses represent errors inherited from \pb~and new errors introduced by \textsc{Ours}, respectively.
The numbers in parentheses represent the quantity of errors, which are initially presented in \pb~and subsequently rectified by \textsc{Ours}.
Type-I means ``semantic discrepancy error''.
Type-II means ``repetition error''.
Type-III means ``morphological error''.}
\label{tab:error_analysis_quantitive}
\end{table}
After conducting the human evaluation in Section~\ref{sec:5.2.1}, we proceed to inspect incorrect instances of \textsc{\pb} and \textsc{Ours} in Zh$\Rightarrow$En test examples.

Furthermore, we summarize incorrect instances into three distinct categories:
(1)~Semantic discrepancy error~(Type-I): The model erroneously suggests irrelevant words.
These words lack semantic relevance to source sentences other than starting with the same human typed characters.
(2)~Repetition error~(Type-II): The model suggests words that convey semantics of source sentences, however, these words already appear within the target context.
(3)~Morphological error~(Type-III): The model suggests incorrect cognates of target words\footnote{It is important to note that some instances might involve valid morphological transformations for the target word, which we do not categorize as errors.}.
In the forthcoming Case Study section, we will present illustrative examples representing each of these three error categories.

In Table~\ref{tab:error_analysis_quantitive}, we present quantitative results of error occurrences for \textsc{\pb} and \textsc{Ours}.
In terms of the total error quantity, \textsc{Ours} exhibits a lower number of errors.
Notably,
for both methods, the most common error type is semantic discrepancy error.
Comparatively,
\textsc{Ours} demonstrates a notable ability to rectify 25 instances~(31.65\%) of Type-I errors, 20 instances~(68.97\%) of Type-II errors and 14 instances~(70.00\%) of Type-III errors that are present in \pb.
Furthermore, \textsc{Ours} exhibits significantly fewer instances in repetition and morphological errors.
However, it is essential to acknowledge that the \textsc{Ours} approach also introduces new incorrect instances in each type that are not originally observed in \textsc{\pb}.

\begin{figure*}
    \centering
    \includegraphics[width=\linewidth]{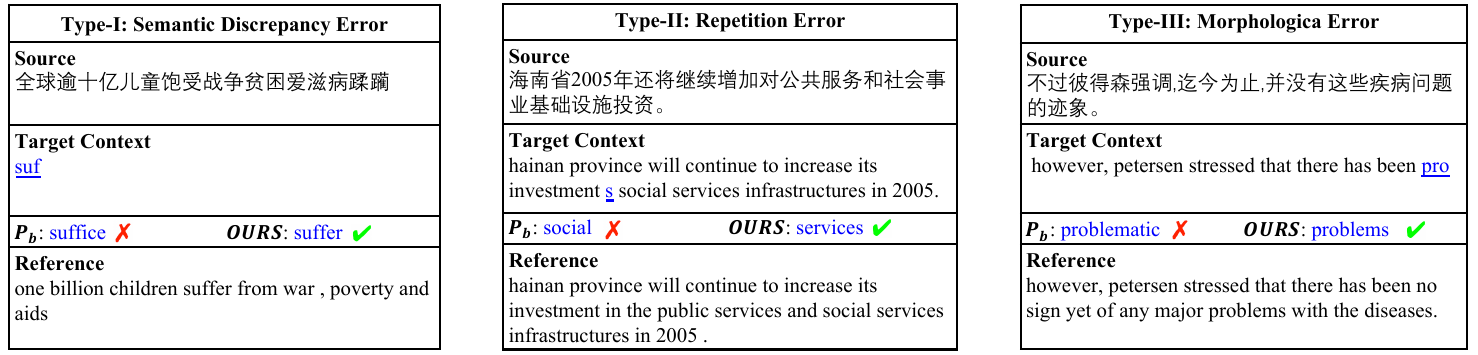}
    \caption{Three cases of \pb \ and \textsc{Ours} in Zh$\Rightarrow$En test set.
Human typed characters are in \underline{underlined} fonts.}
    \label{fig:error_analysis}
\end{figure*}

\begin{figure*}
    \centering
    \includegraphics[width=\linewidth]{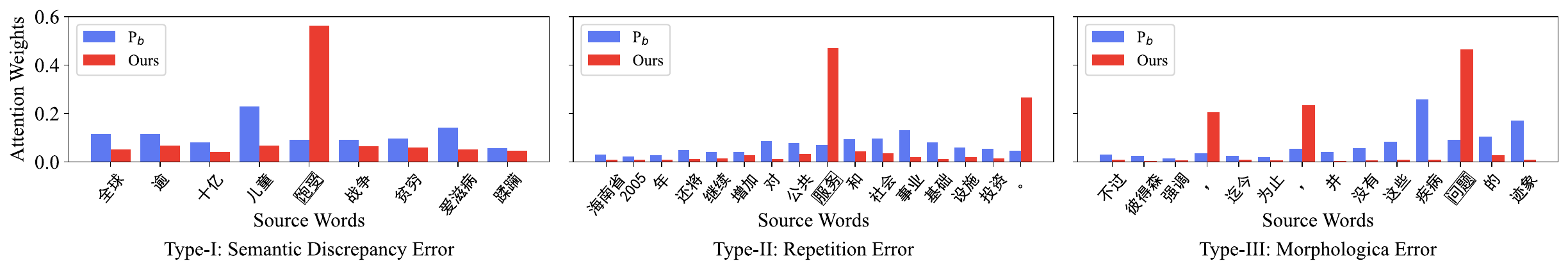}
    \caption{Attention weights from the predicted word to source words of three cases in Figure~\ref{fig:error_analysis}.
    \boxed{\text{Text boxed}} denotes source words aligned with the ground-truth target word.}
    \label{fig:case_study}
\end{figure*}

\paragraph{Case Study}
To better illustrate the advantages of \textsc{Ours} over \textsc{\pb} in utilizing contextual information, thereby leading to enhanced semantic information for word-level autocompletion.
Figure~\ref{fig:error_analysis} presents cases where \textsc{\pb} yields errors while \textsc{Ours} predicts correctly.
Furthermore, Figure~\ref{fig:case_study} illustrates their attention weights which depict the connection between the predicted word and the source words.

In case 1~(Type-I), $P_b$ tends to suggest ``suffice'', which is not consistent with semantics expressed by the source sentence other than starting with human typed characters ``suf''.
In contrast, \textsc{Ours} succeeds in completing ``suf'' to ``suffer''.
Through visualizing attention weights in Figure~\ref{fig:case_study}, we can find that \textsc{Ours} may have the merit of leveraging more information from the valuable source context (e.g., the aligned word ``\begin{CJK}{UTF8}{gbsn}{饱受}\end{CJK}'').
In case 2~(Type-II), $P_b$ completes ``so'' to ``social'', which has already been translated in target context.
With the leverage of interactions between candidate target words and input context, \textsc{Ours} successfully suggests ``services''.
In case 3~(Type-III),
\pb \ suggests the cognates of target words (i.e. ``problematic'').
Whereas, according to the information captured in the energy-based model, \textsc{Ours} succeeds in suggesting the noun ``problems'', which are more appropriate.
Although our model has substantially alleviated aforementioned cases, it is not flawless.
One such instance is that, during the inference stage, the effectiveness of \textsc{Ours} is influenced by the baseline recall rate.\looseness=-1

\paragraph{Running Latency Comparison}
% On the Language Coverage Bias for Neural Machine Translation
\begin{table}[t]
\setlength{\tabcolsep}{3mm}
\centering
\begin{tabular}{lcc}
\toprule
 \multirow{2}{*}{\bf Systems} & \multirow{1}{*}{\bf Training} & \multirow{1}{*}{\bf Inference}\\
  & (hours) & (ms/sample) \\
\midrule
 \textsc{\pb} & 4.19 (1.0$\times$) & 30.01 (1.0$\times$) \\
% 2 & \textsc{Ours} &\quad +4.09 (2.0$\times$) & \quad +16.16 (1.5$\times$) \\
 \textsc{Ours} & 8.28 (2.0$\times$) &  46.17 (1.5$\times$) \\
 \textsc{Trans-BPE} & 4.99 (1.2$\times$) & 56.71 (1.9$\times$) \\ \bottomrule
% 3 & \textsc{Ours}  & 46.17 (1.5$\times$) \\ 
\end{tabular}
\caption{
Training and inference latency comparison on Zh$\Rightarrow$En validation set. ``ms/sample'' represents millisecond per sample.
The evaluation of inference is based on a single NVIDIA V100 GPU, batch size is set to 1, beam size for \textsc{Trans-BPE} is set to 3 and $K$-best size for \textsc{Ours} is 8.
The training latency of \textsc{Ours} does not include the training time of \pb.
}
\label{tab:latency}
\end{table}
Table~\ref{tab:latency} summarizes the training and inference latency of $P_b$, \textsc{Trans-BPE} and \textsc{Ours} on Zh$\Rightarrow$En validation dataset.
The results indicate that the training and inference latency of \textsc{Ours} is comparatively higher than that of \pb~(approximately 2.0 times and 1.5 times, respectively).
This discrepancy in latency can be attributed to the inherent necessity of \textsc{Ours} to get candidate words from $P_b$ and subsequently rerank them, which demands additional computational time.
In comparison to the more potent auto-regressive model, \textsc{Trans-BPE},
\textsc{Ours} exhibits a lower inference latency while concurrently delivering better performance.
As a result, our approach achieves a desirable balance between performance and processing speed.
%thus achieves a desirable trade-off between performance and speed.

\subsection{Applying WLAC into Human-Computer Interactive Translation}
\paragraph{Setup and Evaluation}
As stated in the previous sections, one advantage of WLAC is that it is able to increase the efficiency of human input in interactive machine translation. To exemplify the usefulness of WLAC, we apply the WLAC models into IMT. Specifically, we first implement a practical IMT model following~\citet{DBLP:journals/corr/abs-2105-13072} which is based on lexical constrained decoding~\cite{DBLP:conf/acl/HokampL17} and thus enables the flexible input from users. Then, we apply three WLAC models ($P_b$, \textsc{Trans-BPE} and \textsc{Ours}) into the IMT model, leading to three IMT systems named by IMT-$P_b$, IMT-\textsc{Trans-BPE} and IMT-\textsc{Ours}. As a direct baseline, the IMT system without WLAC is denoted by IMT-\textsc{Raw}.

For efficiency evaluation in IMT, the standard metric, the number of keystrokes from a human translator~\cite{nepveu2004adaptive,bender2005comparison}, is used for all IMT systems. To ensure a fair comparison in efficiency, we enforce all human inputted words to be the same for all IMT systems and thus all these IMT systems yield the same translation outputs. We randomly select a subset consisting of 200 source sentences from Zh$\Rightarrow$En NIST05 as $\boldsymbol{x}$ due to intensive human efforts for in IMT experiments. On this subset, the standard NMT obtains 50.13 BLEU points and all IMT systems achieve 56.02 BLEU points thanks to human interactions. 

% On the Language Coverage Bias for Neural Machine Translation
\begin{table}[t]
\setlength{\tabcolsep}{2mm}
\centering
\begin{tabular}{lccc}
\toprule
\multirow{2}{*}{\textbf{Systems}} & \multirow{2}{*}{\textbf{WLAC}} & \multicolumn{2}{c}{\bf Keystrokes} \\
\cmidrule(lr){3-4}
 & & \bf Total & \bf Average \\
\midrule
 \textsc{IMT-Ours}  & \multirow{3}{*}{\cmark} & 478 & 2.39 \\
\textsc{IMT-Trans-BPE} &  & 686 &  3.43  \\
\textsc{IMT-\pb} & & 704 & 3.52  \\ 
\midrule
 \textsc{IMT-Raw} & \xmark & 1320 & 6.60 \\
\bottomrule
\end{tabular}
\caption{Efficiency for IMT systems with WLAC or not in terms of total and average number of keystrokes. IMT-Raw denotes the IMT system without WLAC function and other systems respectively denote IMT systems with corresponding WLAC models.}
% \yc{Lemao: Please Reorder the first two columns.}
\label{tab:human_imt}
\end{table}

%%%%%%%%%%%%%%%%%%%%%%%%%%%%%%%%%%%%%%%%%%%%%%%%%%%%%%%%%%
% \begin{table*}[t]
%   \resizebox{2.08\columnwidth}{!}{
% \begin{tabular}{ll ccccc ccccc}
% \toprule
%   \multirow{2}{*}{\textbf{\#}} &    \multirow{2}{*}{\textbf{Systems}}         & \multicolumn{5}{c}{\textbf{Zh$\Rightarrow$En}} & \multicolumn{5}{c}{\textbf{De$\Rightarrow$En}} \\ \cmidrule(lr){3-7}\cmidrule(lr){8-12}
%  &            & \multicolumn{1}{c}{\textbf{1}} & \multicolumn{1}{c}{\textbf{2}} & \multicolumn{1}{c}{\textbf{3}} & \multicolumn{1}{c}{\textbf{4}} & \multicolumn{1}{c}{$\bm {\geq}$ \bf{5}} & \multicolumn{1}{c}{\textbf{1}} & \multicolumn{1}{c}{\textbf{2}} & \multicolumn{1}{c}{\textbf{3}} & \multicolumn{1}{c}{\textbf{4}} & \multicolumn{1}{c}{$\bm {\geq}$\bf{5}} \\\midrule
%     1 & \textsc{\pb} &  &  &  &  &  &  &  &  &  &  \\
%     2 & \textsc{Trans-BPE} &  &  &  &  &  &  &  &  &  &  \\
%     3 & \textsc{Ours} & \underline{61.47} & \textbf{68.01} & \textbf{58.47} & \textbf{70.54} & \textbf{64.62} & \textbf{57.17} & \textbf{67.01} & \textbf{56.45} & \textbf{68.28} & \textbf{62.23 }\\ \bottomrule
% \end{tabular}
% }
% \caption{The detailed results for each translation context type of different systems on De$\Leftrightarrow$En validation set.}
% % \caption{The detailed results of different systems on De$\Leftrightarrow$En validation set. ‘$^*$’: results are implemented by ourselves. The best and the second-best results are in \textbf{bold} and \underline{underlined} fonts, respectively.}
% \label{tab:simulation_imt}
% \end{table*}
\begin{figure}
    \centering
    \includegraphics[width=\columnwidth]{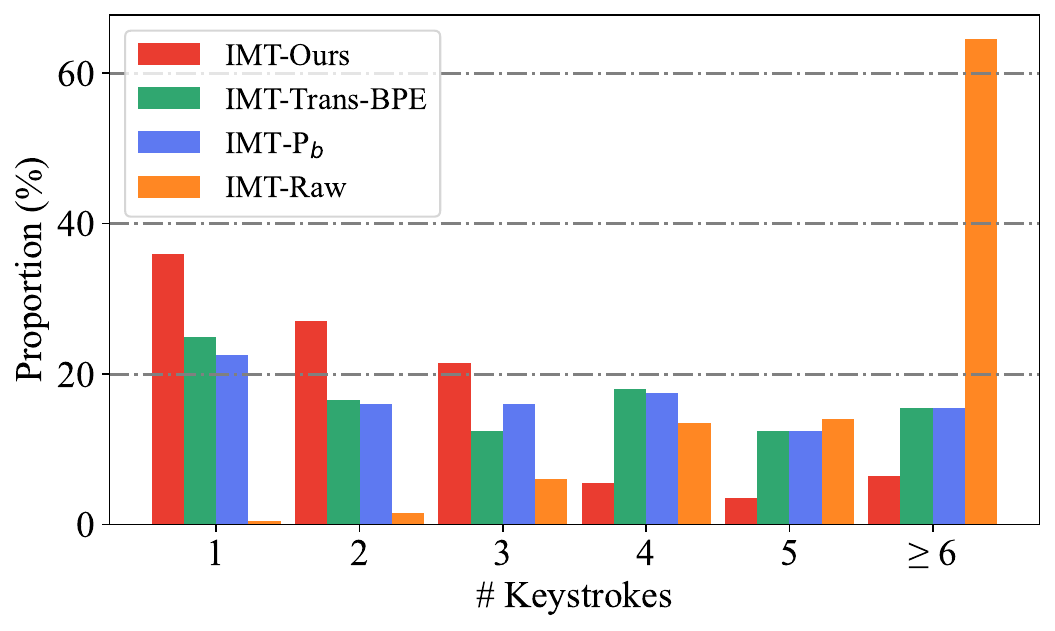}
    \caption{Proportion of the number of keystrokes in different IMT systems with and without WLAC models.}
    \label{fig:keystrokes}
\end{figure}

\paragraph{Experiment Results}
Table~\ref{tab:human_imt} presents the total and average number of keystrokes across different IMT systems.
Notably, the employment of WLAC systems significantly reduces the number of keystrokes in comparison to the IMT-\textsc{Raw} baseline without WLAC.
Furthermore, in comparison to other systems, our proposed IMT-\textsc{Ours} system attains a minimal number of keystrokes relative to other systems.
This observation is reinforced in Figure~\ref{fig:keystrokes}, which depicts the distribution of the number of keystrokes across different systems.
We can see that most of the keystrokes of \textsc{Ours} are less than 3~(constituting approximately 84.5\% of cases), leading to a reduction in the number of keystrokes and offering input convenience for users.\looseness=-1
\section{Related Work}
\paragraph{Computer-aided Translation}
Computer-aided Translation (CAT) \citep{langlais2000transtype,DBLP:journals/coling/BarrachinaBCCCKLNTVV09,DBLP:conf/emnlp/GreenWCHSM14,DBLP:conf/amta/KnowlesK16,DBLP:conf/emnlp/SantyDCB19,DBLP:conf/acl/LeeAPJ21} owns the merit of leveraging advantages of machine translation systems to facilitate human translation process.% \citet{langlais2000transtype} and \citet{DBLP:conf/emnlp/SantyDCB19} propose to predict the next word, which is supposed to be consecutive to the translation prefix.
Word-level AutoCompletion~(WLAC) is an important feature of interactive CAT~\cite{casacuberta2022findings,yang-etal-2022-iigroup} and it plays an important role in CAT. %nevertheless, the research into WLAC remains underexplored.
\citet{DBLP:conf/ijcai/HuangZZZ15} leverage useful source-side knowledge to complete the target word.
\citet{DBLP:conf/acl/LiLHS20} propose a strong word prediction model (WPM) and try to leverage both source-side and target-side information.
However, as stated in Section~\ref{sec:1},
these methods may still inadequately leverage the valuable information from the source sentence.
To fill this gap, we introduce an energy-based model to enable the hidden vector to capture more valuable information.

\paragraph{Reranking}
Reranking has been long researched in natural language processing tasks~\citep{DBLP:conf/naacl/ShenSO04,DBLP:journals/coling/CollinsK05,DBLP:conf/acl/CharniakJ05}.
Recently, the retrieval-then-reranking framework has served as the de facto paradigm \citep{DBLP:journals/corr/abs-1901-04085,DBLP:conf/iclr/ZhangGS0DC22} in text retrieval.
To yield high-quality answers, answer reranking is also widely employed in question answering~\citep{DBLP:conf/iclr/WangY0ZGCWKTC18,DBLP:conf/naacl/IyerMMY21}, dialogue systems~\citep{DBLP:conf/acl/LiYYJLCSJ0Y23} and reasoning~\citep{DBLP:conf/acl/KazemiKBXR23,zhu_2023@core, zhu2023qaap}.
In machine translation,
with the purpose of alleviating the mismatch between maximum likelihood estimation and the desired metric~(e.g., BLEU),
\citet{DBLP:conf/acl/BhattacharyyaRN20} and \citet{DBLP:conf/acl/0001AR20} propose to train an energy-based model to rerank candidate translations generated by NMT models.
In this work, we are in line with prior findings that reranking is a conceptually simple yet empirically powerful framework.
However, we pay more attention to leveraging valuable source sentence information in the WLAC task and corresponding training and inference challenges of the energy-based model for reranking.

\paragraph{Input Method}
In recent years, with the advance of neural networks, the input method has shown significant progress in being effective~\citep{DBLP:conf/acl/HuangLZZ18,DBLP:conf/acl/ZhangHZ19,DBLP:conf/acl/TanDTFHJLS22}.
However, most current research has concentrated on the monolingual scenarios, without sufficient consideration of how to utilize source-side information in bilingual settings~\citep{li2012pinyin, DBLP:conf/ijcai/HuangZZZ15}. 
Our work, which centers on the word-level autocompletion task to reduce keystrokes, is a new exploration of bilingual input methods.
We believe that combining our approach with other input method technologies could significantly enhance the productivity of human translators.
We leave this as a potential direction for future research.

\section{Conclusion}
Word-level AutoCompletion is a critical yet challenging task in Computer-aided Translation. Existing work casts this task as a classification problem.
However, it can not make full use of the contextual information from the input context for its prediction. 
To alleviate such issue, we introduce a reranking perspective by an energy-based model, which directly defines the energy function on top of the input context and the candidate target word.
Extensive experiments and analyses demonstrate the effectiveness of our proposed approach on four standard benchmarks: it achieves about \improvement\% improvements over the strongest baseline.\looseness=-1

%%%%%%%%%%%%%%%%%%%%%%%%%%%%%%%%%%%%%%%%%%%%%%%%%%%%%%%%%%

\section*{Acknowledgments}
This work was partly supported by the National Natural Science Foundation of China (Grant No. U1903213) and the Shenzhen Science and Technology Program (JSGG20220831093004008).
We extend our thanks to annotators for their substantial contributions to this project.
Additionally, we would like to convey our appreciation to the TACL editors and anonymous reviewers for their valuable feedback, which significantly enhanced the paper's quality.

\bibliography{tacl2021}

\begin{thebibliography}{55}
\expandafter\ifx\csname natexlab\endcsname\relax\def\natexlab#1{#1}\fi

\bibitem[{Bahdanau et~al.(2015)Bahdanau, Cho, and Bengio}]{DBLP:journals/corr/BahdanauCB14}
Dzmitry Bahdanau, Kyunghyun Cho, and Yoshua Bengio. 2015.
\newblock \href {http://arxiv.org/abs/1409.0473} {Neural machine translation by jointly learning to align and translate}.
\newblock In \emph{3rd International Conference on Learning Representations, {ICLR} 2015, San Diego, CA, USA, May 7-9, 2015, Conference Track Proceedings}.

\bibitem[{Barrachina et~al.(2009)Barrachina, Bender, Casacuberta, Civera, Cubel, Khadivi, Lagarda, Ney, Tom{\'{a}}s, Vidal, and Vilar}]{DBLP:journals/coling/BarrachinaBCCCKLNTVV09}
Sergio Barrachina, Oliver Bender, Francisco Casacuberta, Jorge Civera, Elsa Cubel, Shahram Khadivi, Antonio~L. Lagarda, Hermann Ney, Jes{\'{u}}s Tom{\'{a}}s, Enrique Vidal, and Juan~Miguel Vilar. 2009.
\newblock \href {https://doi.org/10.1162/coli.2008.07-055-R2-06-29} {Statistical approaches to computer-assisted translation}.
\newblock \emph{Comput. Linguistics}, 35(1):3--28.

\bibitem[{Bender et~al.(2005)Bender, Hasan, Vilar, Zens, and Ney}]{bender2005comparison}
Oliver Bender, Sa{\v{s}}a Hasan, David Vilar, Richard Zens, and Hermann Ney. 2005.
\newblock Comparison of generation strategies for interactive machine translation.
\newblock In \emph{Proceedings of the 10th EAMT Conference: Practical applications of machine translation}.

\bibitem[{Bhattacharyya et~al.(2021)Bhattacharyya, Rooshenas, Naskar, Sun, Iyyer, and McCallum}]{DBLP:conf/acl/BhattacharyyaRN20}
Sumanta Bhattacharyya, Amirmohammad Rooshenas, Subhajit Naskar, Simeng Sun, Mohit Iyyer, and Andrew McCallum. 2021.
\newblock \href {https://doi.org/10.18653/v1/2021.acl-long.349} {Energy-based reranking: Improving neural machine translation using energy-based models}.
\newblock In \emph{Proceedings of the 59th Annual Meeting of the Association for Computational Linguistics and the 11th International Joint Conference on Natural Language Processing, {ACL/IJCNLP} 2021, (Volume 1: Long Papers), Virtual Event, August 1-6, 2021}, pages 4528--4537. Association for Computational Linguistics.

\bibitem[{Casacuberta et~al.(2022)Casacuberta, Foster, Huang, Koehn, Kovacs, Liu, Shi, Watanabe, and Zong}]{casacuberta2022findings}
Francisco Casacuberta, George Foster, Guoping Huang, Philipp Koehn, Geza Kovacs, Lemao Liu, Shuming Shi, Taro Watanabe, and Chengqing Zong. 2022.
\newblock Findings of the word-level autocompletion shared task in wmt 2022.
\newblock In \emph{Proceedings of the Seventh Conference on Machine Translation (WMT)}, pages 812--820.

\bibitem[{Charniak and Johnson(2005)}]{DBLP:conf/acl/CharniakJ05}
Eugene Charniak and Mark Johnson. 2005.
\newblock \href {https://doi.org/10.3115/1219840.1219862} {Coarse-to-fine n-best parsing and maxent discriminative reranking}.
\newblock In \emph{{ACL} 2005, 43rd Annual Meeting of the Association for Computational Linguistics, Proceedings of the Conference, 25-30 June 2005, University of Michigan, {USA}}, pages 173--180. The Association for Computer Linguistics.

\bibitem[{Collins and Koo(2005)}]{DBLP:journals/coling/CollinsK05}
Michael Collins and Terry Koo. 2005.
\newblock \href {https://doi.org/10.1162/0891201053630273} {Discriminative reranking for natural language parsing}.
\newblock \emph{Comput. Linguistics}, 31(1):25--70.

\bibitem[{De~Cao et~al.(2021)De~Cao, Izacard, Riedel, and Petroni}]{DBLP:conf/iclr/CaoI0P21}
Nicola De~Cao, Gautier Izacard, Sebastian Riedel, and Fabio Petroni. 2021.
\newblock \href {https://openreview.net/forum?id=5k8F6UU39V} {Autoregressive entity retrieval}.
\newblock In \emph{9th International Conference on Learning Representations, {ICLR} 2021, Virtual Event, Austria, May 3-7, 2021}. OpenReview.net.

\bibitem[{Devlin et~al.(2019)Devlin, Chang, Lee, and Toutanova}]{DBLP:conf/naacl/DevlinCLT19}
Jacob Devlin, Ming{-}Wei Chang, Kenton Lee, and Kristina Toutanova. 2019.
\newblock \href {https://doi.org/10.18653/v1/n19-1423} {{BERT:} pre-training of deep bidirectional transformers for language understanding}.
\newblock In \emph{Proceedings of the 2019 Conference of the North American Chapter of the Association for Computational Linguistics: Human Language Technologies, {NAACL-HLT} 2019, Minneapolis, MN, USA, June 2-7, 2019, Volume 1 (Long and Short Papers)}, pages 4171--4186. Association for Computational Linguistics.

\bibitem[{Dyer et~al.(2013)Dyer, Chahuneau, and Smith}]{DBLP:conf/naacl/DyerCS13}
Chris Dyer, Victor Chahuneau, and Noah~A. Smith. 2013.
\newblock \href {https://aclanthology.org/N13-1073/} {A simple, fast, and effective reparameterization of {IBM} model 2}.
\newblock In \emph{Human Language Technologies: Conference of the North American Chapter of the Association of Computational Linguistics, Proceedings, June 9-14, 2013, Westin Peachtree Plaza Hotel, Atlanta, Georgia, {USA}}, pages 644--648. The Association for Computational Linguistics.

\bibitem[{Garg et~al.(2019)Garg, Peitz, Nallasamy, and Paulik}]{DBLP:conf/emnlp/GargPNP19}
Sarthak Garg, Stephan Peitz, Udhyakumar Nallasamy, and Matthias Paulik. 2019.
\newblock \href {https://doi.org/10.18653/v1/D19-1453} {Jointly learning to align and translate with transformer models}.
\newblock In \emph{Proceedings of the 2019 Conference on Empirical Methods in Natural Language Processing and the 9th International Joint Conference on Natural Language Processing, {EMNLP-IJCNLP} 2019, Hong Kong, China, November 3-7, 2019}, pages 4452--4461. Association for Computational Linguistics.

\bibitem[{Ghazvininejad et~al.(2019)Ghazvininejad, Levy, Liu, and Zettlemoyer}]{DBLP:conf/emnlp/GhazvininejadLL19}
Marjan Ghazvininejad, Omer Levy, Yinhan Liu, and Luke Zettlemoyer. 2019.
\newblock \href {https://doi.org/10.18653/v1/D19-1633} {Mask-predict: Parallel decoding of conditional masked language models}.
\newblock In \emph{Proceedings of the 2019 Conference on Empirical Methods in Natural Language Processing and the 9th International Joint Conference on Natural Language Processing, {EMNLP-IJCNLP} 2019, Hong Kong, China, November 3-7, 2019}, pages 6111--6120. Association for Computational Linguistics.

\bibitem[{Green et~al.(2014)Green, Wang, Chuang, Heer, Schuster, and Manning}]{DBLP:conf/emnlp/GreenWCHSM14}
Spence Green, Sida~I. Wang, Jason Chuang, Jeffrey Heer, Sebastian Schuster, and Christopher~D. Manning. 2014.
\newblock \href {https://doi.org/10.3115/v1/d14-1130} {Human effort and machine learnability in computer aided translation}.
\newblock In \emph{Proceedings of the 2014 Conference on Empirical Methods in Natural Language Processing, {EMNLP} 2014, October 25-29, 2014, Doha, Qatar, {A} meeting of SIGDAT, a Special Interest Group of the {ACL}}, pages 1225--1236. {ACL}.

\bibitem[{Hokamp and Liu(2017)}]{DBLP:conf/acl/HokampL17}
Chris Hokamp and Qun Liu. 2017.
\newblock \href {https://doi.org/10.18653/v1/P17-1141} {Lexically constrained decoding for sequence generation using grid beam search}.
\newblock In \emph{Proceedings of the 55th Annual Meeting of the Association for Computational Linguistics, {ACL} 2017, Vancouver, Canada, July 30 - August 4, Volume 1: Long Papers}, pages 1535--1546. Association for Computational Linguistics.

\bibitem[{Huang et~al.(2021)Huang, Liu, Wang, Wang, Li, Tu, Huang, and Shi}]{DBLP:journals/corr/abs-2105-13072}
Guoping Huang, Lemao Liu, Xing Wang, Longyue Wang, Huayang Li, Zhaopeng Tu, Chengyan Huang, and Shuming Shi. 2021.
\newblock \href {http://arxiv.org/abs/2105.13072} {Transmart: {A} practical interactive machine translation system}.
\newblock \emph{CoRR}, abs/2105.13072.

\bibitem[{Huang et~al.(2015)Huang, Zhang, Zhou, and Zong}]{DBLP:conf/ijcai/HuangZZZ15}
Guoping Huang, Jiajun Zhang, Yu~Zhou, and Chengqing Zong. 2015.
\newblock \href {http://ijcai.org/Abstract/15/168} {A new input method for human translators: Integrating machine translation effectively and imperceptibly}.
\newblock In \emph{Proceedings of the Twenty-Fourth International Joint Conference on Artificial Intelligence, {IJCAI} 2015, Buenos Aires, Argentina, July 25-31, 2015}, pages 1163--1169. {AAAI} Press.

\bibitem[{Huang et~al.(2018)Huang, Li, Zhang, and Zhao}]{DBLP:conf/acl/HuangLZZ18}
Yafang Huang, Zuchao Li, Zhuosheng Zhang, and Hai Zhao. 2018.
\newblock \href {https://doi.org/10.18653/v1/P18-4024} {Moon {IME:} neural-based chinese pinyin aided input method with customizable association}.
\newblock In \emph{Proceedings of {ACL} 2018, Melbourne, Australia, July 15-20, 2018, System Demonstrations}, pages 140--145. Association for Computational Linguistics.

\bibitem[{Iyer et~al.(2021)Iyer, Min, Mehdad, and Yih}]{DBLP:conf/naacl/IyerMMY21}
Srinivasan Iyer, Sewon Min, Yashar Mehdad, and Wen{-}tau Yih. 2021.
\newblock \href {https://doi.org/10.18653/v1/2021.naacl-main.100} {{RECONSIDER:} improved re-ranking using span-focused cross-attention for open domain question answering}.
\newblock In \emph{Proceedings of the 2021 Conference of the North American Chapter of the Association for Computational Linguistics: Human Language Technologies, {NAACL-HLT} 2021, Online, June 6-11, 2021}, pages 1280--1287. Association for Computational Linguistics.

\bibitem[{Kazemi et~al.(2023)Kazemi, Kim, Bhatia, Xu, and Ramachandran}]{DBLP:conf/acl/KazemiKBXR23}
Mehran Kazemi, Najoung Kim, Deepti Bhatia, Xin Xu, and Deepak Ramachandran. 2023.
\newblock \href {https://doi.org/10.18653/V1/2023.ACL-LONG.361} {{LAMBADA:} backward chaining for automated reasoning in natural language}.
\newblock In \emph{Proceedings of the 61st Annual Meeting of the Association for Computational Linguistics (Volume 1: Long Papers), {ACL} 2023, Toronto, Canada, July 9-14, 2023}, pages 6547--6568. Association for Computational Linguistics.

\bibitem[{Kingma and Ba(2015)}]{DBLP:journals/corr/KingmaB14}
Diederik~P. Kingma and Jimmy Ba. 2015.
\newblock \href {http://arxiv.org/abs/1412.6980} {Adam: {A} method for stochastic optimization}.
\newblock In \emph{3rd International Conference on Learning Representations, {ICLR} 2015, San Diego, CA, USA, May 7-9, 2015, Conference Track Proceedings}.

\bibitem[{Knowles and Koehn(2016)}]{DBLP:conf/amta/KnowlesK16}
Rebecca Knowles and Philipp Koehn. 2016.
\newblock \href {https://aclanthology.org/2016.amta-researchers.9} {Neural interactive translation prediction}.
\newblock In \emph{12th Conferences of the Association for Machine Translation in the Americas: {MT} Researchers' Track, {AMTA} 2016, Austin, TX, USA, October 28 - November 1, 2016}, pages 107--120. The Association for Machine Translation in the Americas.

\bibitem[{Langlais et~al.(2000)Langlais, Foster, and Lapalme}]{langlais2000transtype}
Philippe Langlais, George Foster, and Guy Lapalme. 2000.
\newblock Transtype: a computer-aided translation typing system.
\newblock In \emph{ANLP-NAACL 2000 Workshop: Embedded Machine Translation Systems}.

\bibitem[{LeCun et~al.(2006)LeCun, Chopra, Hadsell, Ranzato, and Huang}]{lecun2006tutorial}
Yann LeCun, Sumit Chopra, Raia Hadsell, M~Ranzato, and F~Huang. 2006.
\newblock A tutorial on energy-based learning.
\newblock \emph{Predicting structured data}, 1(0).

\bibitem[{Lee et~al.(2021{\natexlab{a}})Lee, Auli, and Ranzato}]{DBLP:conf/acl/0001AR20}
Ann Lee, Michael Auli, and Marc'Aurelio Ranzato. 2021{\natexlab{a}}.
\newblock \href {https://doi.org/10.18653/v1/2021.acl-long.563} {Discriminative reranking for neural machine translation}.
\newblock In \emph{Proceedings of the 59th Annual Meeting of the Association for Computational Linguistics and the 11th International Joint Conference on Natural Language Processing, {ACL/IJCNLP} 2021, (Volume 1: Long Papers), Virtual Event, August 1-6, 2021}, pages 7250--7264. Association for Computational Linguistics.

\bibitem[{Lee et~al.(2021{\natexlab{b}})Lee, Ahn, Park, and Jo}]{DBLP:conf/acl/LeeAPJ21}
Dongjun Lee, Junhyeong Ahn, Heesoo Park, and Jaemin Jo. 2021{\natexlab{b}}.
\newblock \href {https://doi.org/10.18653/v1/2021.acl-demo.2} {Intellicat: Intelligent machine translation post-editing with quality estimation and translation suggestion}.
\newblock In \emph{Proceedings of the Joint Conference of the 59th Annual Meeting of the Association for Computational Linguistics and the 11th International Joint Conference on Natural Language Processing, {ACL} 2021 - System Demonstrations, Online, August 1-6, 2021}, pages 11--19. Association for Computational Linguistics.

\bibitem[{Lewis et~al.(2020)Lewis, Liu, Goyal, Ghazvininejad, Mohamed, Levy, Stoyanov, and Zettlemoyer}]{DBLP:conf/acl/LewisLGGMLSZ20}
Mike Lewis, Yinhan Liu, Naman Goyal, Marjan Ghazvininejad, Abdelrahman Mohamed, Omer Levy, Veselin Stoyanov, and Luke Zettlemoyer. 2020.
\newblock \href {https://doi.org/10.18653/v1/2020.acl-main.703} {{BART:} denoising sequence-to-sequence pre-training for natural language generation, translation, and comprehension}.
\newblock In \emph{Proceedings of the 58th Annual Meeting of the Association for Computational Linguistics, {ACL} 2020, Online, July 5-10, 2020}, pages 7871--7880. Association for Computational Linguistics.

\bibitem[{Li(2012)}]{li2012pinyin}
Dong Li. 2012.
\newblock A pinyin input method editor with english-chinese aided translation function.
\newblock In \emph{2012 International Conference on Computer Science and Service System}, pages 446--449. IEEE.

\bibitem[{Li et~al.(2021)Li, Liu, Huang, and Shi}]{DBLP:conf/acl/LiLHS20}
Huayang Li, Lemao Liu, Guoping Huang, and Shuming Shi. 2021.
\newblock \href {https://doi.org/10.18653/v1/2021.acl-long.370} {{GWLAN:} general word-level autocompletion for computer-aided translation}.
\newblock In \emph{Proceedings of the 59th Annual Meeting of the Association for Computational Linguistics and the 11th International Joint Conference on Natural Language Processing, {ACL/IJCNLP} 2021, (Volume 1: Long Papers), Virtual Event, August 1-6, 2021}, pages 4792--4802. Association for Computational Linguistics.

\bibitem[{Li et~al.(2019{\natexlab{a}})Li, Tao, Wu, Feng, Zhao, and Yan}]{DBLP:conf/emnlp/LiTWFZY19}
Jia Li, Chongyang Tao, Wei Wu, Yansong Feng, Dongyan Zhao, and Rui Yan. 2019{\natexlab{a}}.
\newblock \href {https://doi.org/10.18653/v1/D19-1128} {Sampling matters! an empirical study of negative sampling strategies for learning of matching models in retrieval-based dialogue systems}.
\newblock In \emph{Proceedings of the 2019 Conference on Empirical Methods in Natural Language Processing and the 9th International Joint Conference on Natural Language Processing, {EMNLP-IJCNLP} 2019, Hong Kong, China, November 3-7, 2019}, pages 1291--1296. Association for Computational Linguistics.

\bibitem[{Li et~al.(2022)Li, Li, Zhang, Wu, and Liu}]{DBLP:conf/acl/LiLZWL22}
Pengfei Li, Liangyou Li, Meng Zhang, Minghao Wu, and Qun Liu. 2022.
\newblock \href {https://doi.org/10.18653/v1/2022.acl-long.442} {Universal conditional masked language pre-training for neural machine translation}.
\newblock In \emph{Proceedings of the 60th Annual Meeting of the Association for Computational Linguistics (Volume 1: Long Papers), {ACL} 2022, Dublin, Ireland, May 22-27, 2022}, pages 6379--6391. Association for Computational Linguistics.

\bibitem[{Li et~al.(2023{\natexlab{a}})Li, Yang, Yin, Zhu, Cheng, Shang, Jiang, Liu, and Yang}]{li-autoconv}
Siheng Li, Cheng Yang, Yichun Yin, Xinyu Zhu, Zesen Cheng, Lifeng Shang, Xin Jiang, Qun Liu, and Yujiu Yang. 2023{\natexlab{a}}.
\newblock \href {https://doi.org/10.18653/v1/2023.acl-short.149} {{A}uto{C}onv: Automatically generating information-seeking conversations with large language models}.
\newblock In \emph{Proceedings of the 61st Annual Meeting of the Association for Computational Linguistics (Volume 2: Short Papers)}, pages 1751--1762, Toronto, Canada. Association for Computational Linguistics.

\bibitem[{Li et~al.(2023{\natexlab{b}})Li, Yin, Yang, Jiang, Li, Cheng, Shang, Jiang, Liu, and Yang}]{DBLP:conf/acl/LiYYJLCSJ0Y23}
Siheng Li, Yichun Yin, Cheng Yang, Wangjie Jiang, Yiwei Li, Zesen Cheng, Lifeng Shang, Xin Jiang, Qun Liu, and Yujiu Yang. 2023{\natexlab{b}}.
\newblock \href {https://doi.org/10.18653/V1/2023.FINDINGS-ACL.224} {Newsdialogues: Towards proactive news grounded conversation}.
\newblock In \emph{Findings of the Association for Computational Linguistics: {ACL} 2023, Toronto, Canada, July 9-14, 2023}, pages 3634--3649. Association for Computational Linguistics.

\bibitem[{Li et~al.(2019{\natexlab{b}})Li, Li, Liu, Meng, and Shi}]{DBLP:conf/acl/LiLLMS19}
Xintong Li, Guanlin Li, Lemao Liu, Max Meng, and Shuming Shi. 2019{\natexlab{b}}.
\newblock \href {https://doi.org/10.18653/v1/p19-1124} {On the word alignment from neural machine translation}.
\newblock In \emph{Proceedings of the 57th Conference of the Association for Computational Linguistics, {ACL} 2019, Florence, Italy, July 28- August 2, 2019, Volume 1: Long Papers}, pages 1293--1303. Association for Computational Linguistics.

\bibitem[{Li et~al.(2018)Li, Liu, Tu, Shi, and Meng}]{li-etal-2018-target}
Xintong Li, Lemao Liu, Zhaopeng Tu, Shuming Shi, and Max Meng. 2018.
\newblock \href {https://doi.org/10.18653/v1/N18-1125} {Target foresight based attention for neural machine translation}.
\newblock In \emph{Proceedings of the 2018 Conference of the North {A}merican Chapter of the Association for Computational Linguistics: Human Language Technologies, Volume 1 (Long Papers)}, pages 1380--1390, New Orleans, Louisiana. Association for Computational Linguistics.

\bibitem[{Liu et~al.(2016)Liu, Utiyama, Finch, and Sumita}]{liu-etal-2016-neural}
Lemao Liu, Masao Utiyama, Andrew Finch, and Eiichiro Sumita. 2016.
\newblock \href {https://aclanthology.org/C16-1291} {Neural machine translation with supervised attention}.
\newblock In \emph{Proceedings of {COLING} 2016, the 26th International Conference on Computational Linguistics: Technical Papers}, pages 3093--3102, Osaka, Japan. The COLING 2016 Organizing Committee.

\bibitem[{Ma and Collins(2018)}]{DBLP:conf/emnlp/MaC18}
Zhuang Ma and Michael Collins. 2018.
\newblock \href {https://doi.org/10.18653/v1/d18-1405} {Noise contrastive estimation and negative sampling for conditional models: Consistency and statistical efficiency}.
\newblock In \emph{Proceedings of the 2018 Conference on Empirical Methods in Natural Language Processing, Brussels, Belgium, October 31 - November 4, 2018}, pages 3698--3707. Association for Computational Linguistics.

\bibitem[{Nepveu et~al.(2004)Nepveu, Lapalme, Langlais, and Foster}]{nepveu2004adaptive}
Laurent Nepveu, Guy Lapalme, Philippe Langlais, and George Foster. 2004.
\newblock Adaptive language and translation models for interactive machine translation.
\newblock In \emph{Proceedings of the 2004 Conference on Empirical Methods in Natural Language Processing}, pages 190--197.

\bibitem[{Nogueira and Cho(2019)}]{DBLP:journals/corr/abs-1901-04085}
Rodrigo~Frassetto Nogueira and Kyunghyun Cho. 2019.
\newblock \href {http://arxiv.org/abs/1901.04085} {Passage re-ranking with {BERT}}.
\newblock \emph{CoRR}, abs/1901.04085.

\bibitem[{Och and Ney(2003)}]{DBLP:journals/coling/OchN03}
Franz~Josef Och and Hermann Ney. 2003.
\newblock \href {https://doi.org/10.1162/089120103321337421} {A systematic comparison of various statistical alignment models}.
\newblock \emph{Comput. Linguistics}, 29(1):19--51.

\bibitem[{Ott et~al.(2019)Ott, Edunov, Baevski, Fan, Gross, Ng, Grangier, and Auli}]{DBLP:conf/naacl/OttEBFGNGA19}
Myle Ott, Sergey Edunov, Alexei Baevski, Angela Fan, Sam Gross, Nathan Ng, David Grangier, and Michael Auli. 2019.
\newblock \href {https://doi.org/10.18653/v1/n19-4009} {fairseq: {A} fast, extensible toolkit for sequence modeling}.
\newblock In \emph{Proceedings of the 2019 Conference of the North American Chapter of the Association for Computational Linguistics: Human Language Technologies, {NAACL-HLT} 2019, Minneapolis, MN, USA, June 2-7, 2019, Demonstrations}, pages 48--53. Association for Computational Linguistics.

\bibitem[{Ouyang et~al.(2022)Ouyang, Wu, Jiang, Almeida, Wainwright, Mishkin, Zhang, Agarwal, Slama, Ray, Schulman, Hilton, Kelton, Miller, Simens, Askell, Welinder, Christiano, Leike, and Lowe}]{DBLP:conf/nips/Ouyang0JAWMZASR22}
Long Ouyang, Jeffrey Wu, Xu~Jiang, Diogo Almeida, Carroll~L. Wainwright, Pamela Mishkin, Chong Zhang, Sandhini Agarwal, Katarina Slama, Alex Ray, John Schulman, Jacob Hilton, Fraser Kelton, Luke Miller, Maddie Simens, Amanda Askell, Peter Welinder, Paul~F. Christiano, Jan Leike, and Ryan Lowe. 2022.
\newblock \href {http://papers.nips.cc/paper\_files/paper/2022/hash/b1efde53be364a73914f58805a001731-Abstract-Conference.html} {Training language models to follow instructions with human feedback}.
\newblock In \emph{NeurIPS}.

\bibitem[{Ranzato et~al.(2006)Ranzato, Poultney, Chopra, and LeCun}]{DBLP:conf/nips/RanzatoPCL06}
Marc'Aurelio Ranzato, Christopher~S. Poultney, Sumit Chopra, and Yann LeCun. 2006.
\newblock \href {https://proceedings.neurips.cc/paper/2006/hash/87f4d79e36d68c3031ccf6c55e9bbd39-Abstract.html} {Efficient learning of sparse representations with an energy-based model}.
\newblock In \emph{Advances in Neural Information Processing Systems 19, Proceedings of the Twentieth Annual Conference on Neural Information Processing Systems, Vancouver, British Columbia, Canada, December 4-7, 2006}, pages 1137--1144. {MIT} Press.

\bibitem[{Santy et~al.(2019)Santy, Dandapat, Choudhury, and Bali}]{DBLP:conf/emnlp/SantyDCB19}
Sebastin Santy, Sandipan Dandapat, Monojit Choudhury, and Kalika Bali. 2019.
\newblock \href {https://doi.org/10.18653/v1/D19-3018} {{INMT:} interactive neural machine translation prediction}.
\newblock In \emph{Proceedings of the 2019 Conference on Empirical Methods in Natural Language Processing and the 9th International Joint Conference on Natural Language Processing, {EMNLP-IJCNLP} 2019, Hong Kong, China, November 3-7, 2019 - System Demonstrations}, pages 103--108. Association for Computational Linguistics.

\bibitem[{Shen et~al.(2004)Shen, Sarkar, and Och}]{DBLP:conf/naacl/ShenSO04}
Libin Shen, Anoop Sarkar, and Franz~Josef Och. 2004.
\newblock \href {https://aclanthology.org/N04-1023/} {Discriminative reranking for machine translation}.
\newblock In \emph{Human Language Technology Conference of the North American Chapter of the Association for Computational Linguistics, {HLT-NAACL} 2004, Boston, Massachusetts, USA, May 2-7, 2004}, pages 177--184. The Association for Computational Linguistics.

\bibitem[{Shi et~al.(2023)Shi, Su, Yang, Yang, and Cai}]{Shisg}
Chufan Shi, Yixuan Su, Cheng Yang, Yujiu Yang, and Deng Cai. 2023.
\newblock \href {https://doi.org/10.48550/ARXIV.2310.15326} {Specialist or generalist? instruction tuning for specific {NLP} tasks}.
\newblock \emph{CoRR}, abs/2310.15326.

\bibitem[{Tan et~al.(2022)Tan, Dai, Tang, Feng, Huang, Jiang, Li, and Shi}]{DBLP:conf/acl/TanDTFHJLS22}
Minghuan Tan, Yong Dai, Duyu Tang, Zhangyin Feng, Guoping Huang, Jing Jiang, Jiwei Li, and Shuming Shi. 2022.
\newblock \href {https://doi.org/10.18653/v1/2022.acl-long.133} {Exploring and adapting chinese {GPT} to pinyin input method}.
\newblock In \emph{Proceedings of the 60th Annual Meeting of the Association for Computational Linguistics (Volume 1: Long Papers), {ACL} 2022, Dublin, Ireland, May 22-27, 2022}, pages 1899--1909. Association for Computational Linguistics.

\bibitem[{Vaswani et~al.(2017)Vaswani, Shazeer, Parmar, Uszkoreit, Jones, Gomez, Kaiser, and Polosukhin}]{DBLP:conf/nips/VaswaniSPUJGKP17}
Ashish Vaswani, Noam Shazeer, Niki Parmar, Jakob Uszkoreit, Llion Jones, Aidan~N. Gomez, Lukasz Kaiser, and Illia Polosukhin. 2017.
\newblock \href {https://proceedings.neurips.cc/paper/2017/hash/3f5ee243547dee91fbd053c1c4a845aa-Abstract.html} {Attention is all you need}.
\newblock In \emph{Advances in Neural Information Processing Systems 30: Annual Conference on Neural Information Processing Systems 2017, December 4-9, 2017, Long Beach, CA, {USA}}, pages 5998--6008.

\bibitem[{Wang et~al.(2018)Wang, Yu, Jiang, Zhang, Guo, Chang, Wang, Klinger, Tesauro, and Campbell}]{DBLP:conf/iclr/WangY0ZGCWKTC18}
Shuohang Wang, Mo~Yu, Jing Jiang, Wei Zhang, Xiaoxiao Guo, Shiyu Chang, Zhiguo Wang, Tim Klinger, Gerald Tesauro, and Murray Campbell. 2018.
\newblock \href {https://openreview.net/forum?id=rJl3yM-Ab} {Evidence aggregation for answer re-ranking in open-domain question answering}.
\newblock In \emph{6th International Conference on Learning Representations, {ICLR} 2018, Vancouver, BC, Canada, April 30 - May 3, 2018, Conference Track Proceedings}. OpenReview.net.

\bibitem[{Xu et~al.(2022)Xu, Lian, Zhao, Gong, Shou, Jiang, Xie, and Wen}]{DBLP:journals/corr/abs-2206-00212}
Lanling Xu, Jianxun Lian, Wayne~Xin Zhao, Ming Gong, Linjun Shou, Daxin Jiang, Xing Xie, and Ji{-}Rong Wen. 2022.
\newblock \href {https://doi.org/10.48550/arXiv.2206.00212} {Negative sampling for contrastive representation learning: {A} review}.
\newblock \emph{CoRR}, abs/2206.00212.

\bibitem[{Yang et~al.(2022{\natexlab{a}})Yang, Li, Shi, and Yang}]{yang-etal-2022-iigroup}
Cheng Yang, Siheng Li, Chufan Shi, and Yujiu Yang. 2022{\natexlab{a}}.
\newblock \href {https://aclanthology.org/2022.wmt-1.121} {{IIGROUP} submissions for {WMT}22 word-level {A}uto{C}ompletion task}.
\newblock In \emph{Proceedings of the Seventh Conference on Machine Translation (WMT)}, pages 1187--1191, Abu Dhabi, United Arab Emirates (Hybrid). Association for Computational Linguistics.

\bibitem[{Yang et~al.(2022{\natexlab{b}})Yang, Meng, Zhang, Li, and Zhou}]{DBLP:conf/emnlp/YangMZL022}
Zhen Yang, Fandong Meng, Yingxue Zhang, Ernan Li, and Jie Zhou. 2022{\natexlab{b}}.
\newblock \href {https://doi.org/10.18653/V1/2022.EMNLP-MAIN.353} {Wets: {A} benchmark for translation suggestion}.
\newblock In \emph{Proceedings of the 2022 Conference on Empirical Methods in Natural Language Processing, {EMNLP} 2022, Abu Dhabi, United Arab Emirates, December 7-11, 2022}, pages 5278--5290. Association for Computational Linguistics.

\bibitem[{Zhang et~al.(2022)Zhang, Gong, Shen, Lv, Duan, and Chen}]{DBLP:conf/iclr/ZhangGS0DC22}
Hang Zhang, Yeyun Gong, Yelong Shen, Jiancheng Lv, Nan Duan, and Weizhu Chen. 2022.
\newblock \href {https://openreview.net/forum?id=MR7XubKUFB} {Adversarial retriever-ranker for dense text retrieval}.
\newblock In \emph{The Tenth International Conference on Learning Representations, {ICLR} 2022, Virtual Event, April 25-29, 2022}. OpenReview.net.

\bibitem[{Zhang et~al.(2019)Zhang, Huang, and Zhao}]{DBLP:conf/acl/ZhangHZ19}
Zhuosheng Zhang, Yafang Huang, and Hai Zhao. 2019.
\newblock \href {https://doi.org/10.18653/v1/p19-1154} {Open vocabulary learning for neural chinese pinyin {IME}}.
\newblock In \emph{Proceedings of the 57th Conference of the Association for Computational Linguistics, {ACL} 2019, Florence, Italy, July 28- August 2, 2019, Volume 1: Long Papers}, pages 1584--1594. Association for Computational Linguistics.

\bibitem[{Zhu et~al.(2023{\natexlab{a}})Zhu, Wang, Zhang, Zhang, Huang, Gan, Zhang, and Yang}]{zhu_2023@core}
Xinyu Zhu, Junjie Wang, Lin Zhang, Yuxiang Zhang, Yongfeng Huang, Ruyi Gan, Jiaxing Zhang, and Yujiu Yang. 2023{\natexlab{a}}.
\newblock \href {https://doi.org/10.18653/v1/2023.acl-long.245} {Solving math word problems via cooperative reasoning induced language models}.
\newblock In \emph{Proceedings of the 61st Annual Meeting of the Association for Computational Linguistics (Volume 1: Long Papers)}, pages 4471--4485, Toronto, Canada. Association for Computational Linguistics.

\bibitem[{Zhu et~al.(2023{\natexlab{b}})Zhu, Yang, Chen, Li, Lou, and Yang}]{zhu2023qaap}
Xinyu Zhu, Cheng Yang, Bei Chen, Siheng Li, Jian{-}Guang Lou, and Yujiu Yang. 2023{\natexlab{b}}.
\newblock \href {https://doi.org/10.48550/ARXIV.2305.14221} {Question answering as programming for solving time-sensitive questions}.
\newblock \emph{CoRR}, abs/2305.14221.

\end{thebibliography}
\bibliographystyle{acl_natbib}

\end{document}